\documentclass[Afour,sageh,times]{sagej}

\usepackage{moreverb,url}
\usepackage{bm}
\usepackage{algorithm}
\usepackage[noend]{algpseudocode}
\usepackage{mathrsfs}
\usepackage{booktabs}       % professional-quality tables
\usepackage{amsfonts}       % blackboard math symbols
\usepackage{nicefrac}       % compact symbols for 1/2, etc.
\usepackage{microtype}      % microtypography
\usepackage{graphicx}
\usepackage{subcaption}
\usepackage{mathtools}
\usepackage{amsmath}

\usepackage[colorlinks,bookmarksopen,bookmarksnumbered,citecolor=red,urlcolor=red]{hyperref}

\newcommand\BibTeX{{\rmfamily B\kern-.05em \textsc{i\kern-.025em b}\kern-.08em
T\kern-.1667em\lower.7ex\hbox{E}\kern-.125emX}}

\def\volumeyear{2019}

\usepackage{amsmath, amssymb, amsthm}
\usepackage{enumerate}
\usepackage{tikz}
\usepackage{multirow}
\usetikzlibrary{shapes.geometric}
\newcommand{\myignore}[1]{}

\newcommand{\inparen}[1]{\left ( #1 \right )}

\def \F {\mathbf{F}}
\def \G {\mathbf{G}}

\def \U {\mathbf{U}}

\begin{document}

\graphicspath{{./Graphics/}}

\runninghead{Shah et al.}

\title{Supervised Bayesian Specification Inference from Demonstrations}

\author{Ankit Shah\affilnum{1}, Pritish Kamath\affilnum{1}, Shen Li\affilnum{1}, Patrick Craven\affilnum{2}, Kevin Landers\affilnum{2}, Kevin Oden\affilnum{2} and Julie Shah\affilnum{1}}

\affiliation{\affilnum{1}CSAIL, MIT\\
\affilnum{2}Lockheed Martin Corporation}

\corrauth{Ankit Shah
Computer Science and Artificial Intelligence Laboratory,
Massachusetts Institute of Technology,
Cambrdige,
MA 02139,
USA.}

\email{ajshah@mit.edu}

\begin{abstract}
When observing task demonstrations, human apprentices are able to identify whether a given task is executed correctly long before they gain expertise in actually performing that task. Prior research into learning from demonstrations (LfD) has failed to capture this notion of the acceptability of a task’s execution; meanwhile, temporal logics provide a flexible language for expressing task specifications. Inspired by this, we present Bayesian specification inference, a probabilistic model for inferring task specification as a temporal logic formula. We incorporate methods from probabilistic programming to define our priors, along with a domain-independent likelihood function to enable sampling-based inference. We demonstrate the efficacy of our model for inferring specifications, with over 90\% similarity observed between the inferred specification and the ground truth -- both within a synthetic domain and during a real-world table setting task.
\end{abstract}


\maketitle

\section{Introduction}
\label{Sec:Intro}
Imagine showing a friend how to play your favorite quest-based video game. A mission within such a game might be composed of multiple sub-quests that must be completed in order to finish that level. In this scenario, it is likely that your friend would comprehend what needs to be done in order to complete the mission well before he or she was actually able to play the game effectively. While learning from demonstrations, human apprentices can identify whether a task is executed correctly long before gaining expertise in that task. In the context of learning from demonstrations for robotic tasks, a system that can evaluate the acceptability of an execution before learning to execute a task can lead to more-focused exploration of execution strategies. Further, a system that can express its specifications would be more transparent with regard to its objectives than a system that simply imitates the demonstrator. Such characteristics are highly desirable in applications such as manufacturing or disaster response, where the cost of a mistake can be especially high. Finally, a robotic system with a correct understanding of the acceptability of executions can explore more-creative execution traces not present in the demonstrated set.

Most current approaches to learning from demonstration frame this problem as one of learning a reward function or policy within the setting of a Markov decision process; however, user specification of acceptable behaviors through reward functions and policies remains an open problem \cite{arnold2017value}. Temporal logics have been used in prior research as a language for expressing desirable system behaviors, and can improve the interpretability of specifications if expressed as compositions of simpler templates (akin to those described by \cite{dwyer1999patterns}). In this work, we propose a probabilistic model for inferring a task’s temporal structure as a linear temporal logic (LTL) specification.

A specification inferred from demonstrations is valuable in conjunction with synthesis algorithms for verifiable controllers (\cite{kress2009temporal} and \cite{raman2015reactive}), as a reward signal during reinforcement learning (\cite{li2017reinforcement} and \cite{littman2017environment}), and as a system model for execution monitoring. In our work, we frame specification learning as a Bayesian inference problem.

The flexibility of LTL for specifying behaviors also represents a key challenge with regard to inference due to a large hypothesis space. We define prior and likelihood distributions over a smaller but relevant part of the LTL formulas, using templates based on work by \cite{dwyer1999patterns}. Ideas from universal probabilistic programming languages formalized by \cite{FreerRT14} and \cite{goodman2012church,dippl} are key to our modeling approach; indeed, probabilistic programming languages enabled \cite{ellis2017learning,ellis2015unsupervised} to perform inference over complex, recursively defined hypothesis spaces of graphics programs and pronunciation rules.

We evaluate our model’s performance within three domains. First, we incorporate a synthetic domain and a real-world task involving setting a dinner table, both of which are representative of candidate tasks for robots to learn from demonstrations. For both these domains, the ground-truth specifications are known, and we report the capability of our model to achieve greater than 90\% similarity between the inferred and ground-truth specifications. We also demonstrate the capability of our model to infer mission objective specifications for evaluating large-force combat flying exercises involving multiple friendly and hostile aircrafts. The LFE domain is particularly challenging, as it incorporates multiple decision-making participants, some of which act cooperatively and some in an adversarial fashion. We demonstrate that our model makes predictions that are well-aligned with those of an expert acting as the commander for an example LFE mission. Further, we demonstrate that our method of using template compositions allows for an interpretable decision boundary for the classifier inferred by our model.

Bayesian specification inference was first introduced in work by \cite{shah2018nips}; in this paper, we present an advancement of that work and apply the model to newer evaluation domains. First, we extend the probabilistic model to be capable of learning both inductively (from positive examples only) and from positive and negative examples. We also extend the evaluation presented by \cite{shah2018nips} to include the large-force exercise domain.

% we extend the probabilistic model to be framed as a supervised learning problem capable of learning either purely inductively (from positive examples only) or with positive and negative examples akin to supervised learning. We also

%$We demonstrate the capability of our model to achieve greater than 90\% similarity between the ground truth specification and the inferred specification, both within a synthetic domain and a real-world task of setting a dinner table.  Finally we demonstrate the capability of our model to infer interpretable decision boundaries on a supervised learning task of evaluating the execution of large force exercises for combat flying.

\section{Related Work}
\label{Sec:related}

One common approach discussed in prior research frames learning from demonstration as an inverse reinforcement learning (IRL) problem. \cite{ng2000algorithms} and \cite{abbeel2004apprenticeship} first formalized the problem of inverse reinforcement learning as one of optimization in order to identify the reward function that best explains observed demonstrations. \cite{ziebart2008maximum} introduced algorithms to compute optimal policy for imitation using the maximum entropy criterion. \cite{konidaris2012robot} and \cite{niekum2015learning} framed IRL in a semi-Markov setting, allowing for implicit representation of the temporal structure of the task. Surveys by \cite{argall2009survey} and \cite{chernova2014robot} provided a comprehensive review of techniques built upon these works as applied to robotics. However, according to \cite{arnold2017value}, one drawback of inverse reinforcement learning is the non-triviality of extracting task specifications from a learned reward function or policy. Our method bridges this gap by directly learning the specifications for acceptable execution of a given task.

%Hayes and Scassellati \cite{hayes2016icra} proposed an algorithm to learn a hierarchical task network based representation from demonstration.

Temporal logics, introduced by \cite{pnueli1977temporal}, are an expressive grammar used to describe the desirable temporal properties of task execution. Temporal logics have previously served as a language for goal definitions in reinforcement learning algorithms (\cite{li2017reinforcement} and \cite{littman2017environment}), reactive controller synthesis (\cite{kress2009temporal}) and \cite{raman2015reactive}), and domain-independent planning (\cite{kim2017collaborative}).

\cite{kasenberg2017interpretable} explored mining globally persistent specifications from optimal traces of a finite-state Markov decision process (MDP). \cite{jin2015mining} proposed algorithms for mining temporal specifications similar to rise and setting times for closed-loop control systems. Works by \cite{kong2014temporal}, \cite{kong2017temporal}, \cite{yoo2017rich}, and \cite{lemieux2015general} are most closely related to our own, as our work incorporates only the observed state variable (and not the actions of the demonstrators) as input to the model. \cite{lemieux2015general} introduced Texada, a general-specification mining tool for software logs. Texada outputs all possible satisfied instances of a particular formula template; however, it treats each time step as a string, with all unique strings within the log treated as unique propositions. Texada would treat a system with $n$ propositions as a system with $2^n$ distinct propositions; thus, interpreting a mined formula is non-trivial. \cite{kong2014temporal}, \cite{kong2017temporal}, and \cite{yoo2017rich} mined PSTL specifications for given demonstrations while simultaneously inferring signal propositions akin to our own user-defined atomic propositions by conducting breadth-first search over a directed acyclic graph (DAG formed by candidate formulas. Our prior specifications allow for better connectivity between different formulas, while using MCMC-based approximate inference enables fixed runtimes.

%Their method requires searching through a DAG over formulas induced by a partial ordering relation between formulas. The size of the DAG increases exponentially with the number of signal predicates to be included in the formula. Furthermore, to reach a formula of size $n$, their algorithm must search through all formulas smaller than $n$

We adopt a fully Bayesian approach to model the inference problem, allowing our model to maintain a posterior distribution over candidate formulas. This distribution provides a measure of confidence when predicting the acceptability of a new demonstration that the aforementioned approaches do not.

\section{Linear Temporal Logic}
\label{Sec:prelims}

Linear temporal logic (LTL), introduced by \cite{pnueli1977temporal}, provides an expressive grammar for describing temporal behaviors. A LTL formula is composed of atomic propositions (discrete time sequences of Boolean literals) and both logical and temporal operators, and is interpreted over traces $[\bm{\alpha}]$ of the set of propositions, $\bm{\alpha}$. The notation $[\bm{\alpha}],t\models \varphi$ indicates that $\varphi$ holds at time $t$. The trace $[\bm{\alpha}]$ satisfies $\varphi$ (denoted as $[\bm{\alpha}]\models \varphi$) iff $[\bm{\alpha}],0\models \varphi$. The minimal syntax of LTL can be described as follows:

\begin{equation}
  \varphi::= p \mid \neg\varphi_1 \mid \varphi_1\vee\varphi_2 \mid \mathbf{X}\varphi_1 \mid \varphi_1\mathbf{U}\varphi_2
\end{equation}

$p$ is an atomic proposition; $\varphi_1$ and $\varphi_2$ are valid LTL formulas. The operator $\mathbf{X}$ is read as `next' and $\mathbf{X}\varphi_1$ evaluates as true at time $t$ if $\varphi_1$ evaluates to true at $t+1$. The operator $\mathbf{U}$ is read as `until' and the formula $\varphi_1 \mathbf{U} \varphi_2$ evaluates as true at time $t_1$ if $\varphi_2$ evaluates as true at some time $t_2 > t_1$ and $\varphi_1$ evaluates as true for all time steps $t$ such that $t_1\leq t\leq t_2$. In addition to the minimal syntax, we also use the additional first- order logic operators $\wedge$ (and) and $\mapsto$ (implies), as well as other higher-order temporal operators, $\mathbf{F}$ (eventually) and $\mathbf{G}$ (globally). $\mathbf{F}\varphi_1$ evaluates to true at $t_1$ if $\varphi_1$ evaluates as true for some $t\geq t_1$. $\mathbf{G}\varphi_1$ evaluates to true at $t_1$ if $\varphi_1$ evaluates as true for all $t\geq t_1$.

\section{Bayesian Specification Inference}
\label{Sec:method}

A large number of tasks comprised of multiple subtasks can be represented by a combination of three temporal behaviors among those defined by \cite{dwyer1999patterns} — namely, global satisfaction of a proposition, eventual completion of a subtask, and temporal ordering between subtasks. With $\varphi_{global}$, $\varphi_{eventual}$, and $\varphi_{order}$ representing LTL formulas for these behaviors, the task specification is written as follows:
\vspace{-5pt}
\begin{equation}
  \varphi = \varphi_{global} \wedge \varphi_{eventual} \wedge \varphi_{order}
  \label{Eq:Template}
\end{equation}

We represent the task demonstrations as an observed sequence of state variables, $[\bm{x}]$. Let $\bm{\alpha} in \{0,1\}^n$ represent a vector of finite dimension formed by $n$ Boolean propositions. The propositions are related to the state variables through a labeling function, $\bm{\alpha} = f(\bm{x})$, which is known a priori.

The inference model is provided a label, $y$, to indicate whether an execution is acceptable or not, along with the actual demonstrations. Thus, the training set $\bm{D} = \{ ( [\bm{\alpha}]_i, y_i ) ~;~ i\in \{1,2,\hdots,n_{demo}\} \}$ consists of $n_{demo}$ demonstrations along with the label. The output, again, is a probability distribution $P(\varphi | \bm{D})$.

\subsection{Formula Template}
\label{SS:FormulaTemplate}

\textbf{Global satisfaction}: Let $\mathbf{T}$ be the set of candidate propositions to be globally satisfied, and let $\bm{\tau} \subseteq \bm{T}$ be the actual subset of satisfied propositions. The LTL formula that specifies this behavior is written as follows:
\begin{equation}
  \varphi_{global} =
  \inparen{ \bigwedge_{\tau \in \bm{\tau}} \inparen{ \G(\tau) }}
  \label{Eq:Global}
\end{equation}

Such formulas are useful for specifying that some constraints must always be met -- for example, a robot must avoid collisions while in motion, or an aircraft must avoid no-fly zones.

\textbf{Eventual completion}: Let $\bm{\Omega}$ be the set of all candidate subtasks, and let $\bm{W_1}\subseteq \bm{\Omega}$ be the set of subtasks that must be completed if the conditions represented by $\pi_w; w\in \bm{W_1}$ are met. $\omega_w$ are propositions representing the completion of a subtask. The LTL formula that specifies this behavior is written as follows:

\begin{equation}
  \varphi_{eventual} =
  \inparen{ \bigwedge_{w\in \mathbf{W_1}} \inparen{ \pi_w \rightarrow \F\omega_w } }
  \label{Eq:Eventually}
\end{equation}

\textbf{Temporal ordering}: Every set of feasible ordering constraints over a set of subtasks is mapped to a DAG over nodes representing these subtasks. Each edge in the DAG corresponds to a binary precedence constraint. Let $\bm{W_2}$ be the set of binary temporal orders defined by $\bm{W_2} = \{ (w_1,w_2) : w_1\in\bm{V}, w_2\in \text{Descendants}(w_1) \}$, where $\bm{V}$ is the set of all nodes within the task graph. Thus, the ordering constraints include an enumeration of not just the edges in the task-graph, but all descendants of a given node. For subtasks $w_1$ and $w_2$, the ordering constraint is written as follows:

\begin{equation}
  \varphi_{order} = \inparen{ \bigwedge_{(w_1, w_2) \in \mathbf{W_2}} \inparen{ \pi_{w_1} \rightarrow (\neg \omega_{w_2} \U \omega_{w_1})}}
  \label{Eq:Order}
\end{equation}

This formula states that if conditions for the execution of $w_1$ i.e. $\pi_{w_1}$ are satisfied, $w_2$ must not be completed until $w_1$ has been completed.

For the purposes of this paper, we assume that all required propositions $\bm{\alpha} = [\bm{\tau}, \bm{\pi}, \bm{\omega}]^T$ and labeling functions $f(\bm{x})$ are known, along with the sets $\bm{T}$ and $\bm{\Omega}$ and the mapping of the condition propositions $\pi_w$ to their subtasks. Given these assumptions, the problem of inferring the correct formula for a task is equivalent to identifying the correct subsets $\bm{\tau}$, $\bm{W_1}$, and $\bm{W_2}$, that explain the observed demonstrations well.

\subsection{Specification Learning as Bayesian Inference}
\label{SS:Bayesian formulation}
The Bayes theorem is fundamental to the problem of inference, and is stated as follows:
\begin{equation}
  P(h\mid \mathbf{D}) = \frac{P(h)P(\mathbf{D} \mid h)}{\sum_{h\in\mathbf{H}} P(h) P(\mathbf{D} \mid h)}
  \label{Eq:Bayes}
\end{equation}
$P(h)$ is the prior distribution over the hypothesis space, and $P(\mathbf{D} \mid h)$ is the likelihood of observing the data given a hypothesis. Our hypothesis space is defined by $\bm{H} = \bm{\varphi}$, where $\bm{\varphi}$ is the set of all formulas that can be generated by the production rule defined by the template in Equation \ref{Eq:Template}. The observed data comprises the set of demonstrations provided to the system by expert demonstrators (note that we assume all these demonstrations are acceptable). $\bm{D}$ is the training dataset.

\subsubsection{Prior specification}
\label{SS:Priors}

\begin{table*}
  \footnotesize
  \caption{Prior definitions and hyperparameters.}
  \label{Tab:priors}
  \centering
  \begin{tabular}{lll}
    \toprule
    Prior & $\varphi_{Order}$ & Hyperparameters                   \\
    \midrule

    Prior 1  & \texttt{RandomPermutation}($\bm{\Omega}$)                    & $p_G, p_E$  \\
    Prior 2  & \texttt{SampleSetsOfLinearChains}($\bm{\Omega},p_{part}$)   & $p_G, p_E, p_{part}$  \\
    Prior 3  & \texttt{SampleForestofSubTasks}($\bm{\Omega},N_{new}$)      & $p_G, p_E, N_{new}$  \\
    \bottomrule
  \end{tabular}
\end{table*}

While sampling candidate formulas as per the template depicted in Equation \ref{Eq:Template}, we treat the sub-formulas in Equations \ref{Eq:Global}, \ref{Eq:Eventually}, and \ref{Eq:Order} as independent to each other. As generating the actual formula, given the selected subsets, is deterministic, sampling $\varphi_{global}$ and $\varphi_{eventual}$ is equivalent to selecting a subset of a given finite universal set. Given a set $A$, we define \texttt{SampleSubset}($A$,$p$) as the process of applying a Bernoulli trial with a success probability of $p$ to each element of A and returning the subset of elements for which the trial was successful. Thus, sampling $\varphi_{global}$ and $\varphi_{eventual}$ is accomplished by performing \texttt{SampleSubset}($\bm{T},p_{G})$ and \texttt{SampleSubset}$(\bm{\Omega},p_{E})$. Sampling $\varphi_{order}$ is equivalent to sampling a DAG, with the nodes of the graph representing subtasks. Based on domain knowledge, appropriately constraining the DAG topologies would result in better inference with fewer demonstrations. Here, we present three possible methods of sampling a DAG, with different restrictions on the graph topology.

\begin{algorithm}
\scriptsize
\begin{algorithmic}[1]
\Function{SampleSetsOfLinearChain}{$\bm{\Omega}$,$p_{part}$}
\State $i\gets1$; $\bm{C_i} \gets []$ \label{lin:init}
\State $\bm{P} \gets$ random permutation($\bm{\Omega}$) \label{lin:rp_lc}
\For{$a\in \bm{P}$}
\State $\bm{C_i}$.append($a$) \label{lin:add2chain}
\State $k \gets$ Bernoulli($p_{part}$)\label{lin:bern}
\If {$k=1$}
\State $i=i+1$; $\bm{C_i} \gets []$ \label{lin:newchain}
\EndIf
\EndFor
\State \Return $\bm{C_j}~ \forall~ j$
\EndFunction
\end{algorithmic}
\caption{SampleSetsOfLinearChains}
\label{Alg:SetLC}
%\vspace{-15}
\end{algorithm}

{\bf{Linear chains:}}
A linear chain is a DAG such that all subtasks must occur within a single, unique sequence out of all permutations. Sampling a linear chain is equivalent to selecting a permutation from a uniform distribution, and is achieved via the following probabilistic program: for a set of size $n$, sample $n-1$ elements from that set without replacement, with uniform probability.

{\bf{Sets of linear chains:}}
This graph topology includes graphs formed by a set of disjoint sub-graphs, each of which is either a linear chain or a solitary node. The execution of subtasks within a particular linear chain must be completed in the specified order; however, no temporal constraints exist between the chains. Algorithm \ref{Alg:SetLC} depicts a probabilistic program for constructing these sets of chains. In line \ref{lin:init}, the first active linear chain is initialized as an empty sequence. In line \ref{lin:rp_lc}, a random permutation of the nodes is produced. For each element $a \in \bm{P}$, line \ref{lin:add2chain} adds the element to the last active chain. Lines \ref{lin:bern} and \ref{lin:newchain} ensure that after each element, either a new active chain is initiated (with a probability of $p_{part}$) or the old active chain continues (with a probability of $1-p_{part}$).

{\bf{Forest of sub-tasks:}}
This graph topology includes forests (i.e., sets of disjoint trees). A given node has no temporal constraints with respect to its siblings, but must precede all its descendants. Algorithm \ref{Alg:Forest} depicts a probabilistic program for sampling a forest. Line \ref{lin:rp_forest} creates $\bm{P}$, a random permutation of the subtasks. Line \ref{lin:initForest} initializes an empty forest. In order to support a recursive sampling algorithm, the data structure representing forests is defined as an array of trees, $\bm{\mathcal{F}}$. The $i^{th}$ tree has two attributes: a root node, $\bm{\mathcal{F}}[i].\text{root}$, and a `descendant forest,' $\bm{\mathcal{F}}[i].\text{descendant}$, in which the root node of each tree is a child of the root node defined as the first attribute. The length of the forest, $\bm{\mathcal{F}}.\text{length}$, is the number of trees included in that forest. The size of a tree, $\bm{\mathcal{F}}[i].\text{size}$, is the number of nodes within the tree (i.e., the root node and all of its descendants). For each subtask in the random permutation $\bm{P}$, line \ref{lin:InsertElement} inserts the given subtask into the forest as per the recursive function \texttt{InsertIntoForest} defined in lines \ref{lin:BeginIIF} through \ref{lin:EndIIF}. In line \ref{lin:Sample}, an integer $i$ is sampled from a categorical distribution, with $\{1,2,\ldots,\bm{\mathcal{F}}.\text{length}+1\}$ as the possible outcomes. The probability of each outcome is proportional to the size of the trees in the forest, while the probability of $\bm{\mathcal{F}}.\text{length}+1$ being the outcome is proportional to $N_{new}$, a user-defined parameter. This sampling process is similar in spirit to the Chinese restaurant process (\cite{crp}). If the outcome of the draw is $\bm{\mathcal{F}}.\text{length}+1$, then a new tree with root node $a$ is created in line \ref{lin:CreateNew}; otherwise, \texttt{InsertIntoForest} is called recursively to add $a$ to the forest $\bm{\mathcal{F}}[i].\text{descendants}$, as per line \ref{lin:AddRecursive}.

%\vspace{-15}
\begin{algorithm}
\scriptsize
\begin{algorithmic}[1]
\Function{SampleForestofSubtasks}{$\bm{\Omega}$,$N_{new}$}

\State $\bm{P} \gets$ random permutation($\bm{\Omega}$) \label{lin:rp_forest}
\State $\bm{\mathcal{F}}\gets[]$ \label{lin:initForest}
\For{$a \in \bm{P}$}
\State $\bm{\mathcal{F}} = $InsertIntoForest($\bm{\mathcal{F}}$,$a$) \label{lin:InsertElement}
\EndFor
\State \Return $\bm{\mathcal{F}}$
\EndFunction

\Function{InsertIntoForest}{$\bm{\mathcal{F}}$, $a$} \label{lin:BeginIIF}
\State $i \gets$ Categorical$([\bm{\mathcal{F}}[1].\text{size}, \bm{\mathcal{F}}[2].\text{size}, \ldots, \bm{\mathcal{F}}[\bm{\mathcal{F}}.\text{length}].\text{size}, N_{new} ])$ \label{lin:Sample}
\If{$i = \bm{\mathcal{F}}.\text{length}+1$}
\State Create new tree $\bm{\mathcal{F}}[\bm{\mathcal{F}}.\text{length}+1].\text{root}=a$  \label{lin:CreateNew}
\Else
\State $\bm{\mathcal{F}}[i].\text{descendants}$ = InsertIntoForest($\bm{\mathcal{F}}[i].\text{descendants}$, $a$) \label{lin:AddRecursive}
\EndIf
\State \Return $\bm{\mathcal{F}}$
\EndFunction\label{lin:EndIIF}

\end{algorithmic}
%\vspace{-15}
\caption{SampleForestofSubtasks}
\label{Alg:Forest}
%\vspace{-15}
\end{algorithm}
%\vspace{-20}

Three prior distributions based on the four probabilistic programs are described in \autoref{Tab:priors}. In all the priors, $\varphi_{global}$ and $\varphi_{eventual}$ are sampled using \texttt{SampleSubset}($\bm{T}, p_G$) and \texttt{SampleSubset}($\bm{\Omega}, p_E$), respectively.

\subsubsection{Likelihood function}
\label{SS:Likelihood}

The likelihood distribution, $P(\{ \left[ \bm{\alpha} \right]_i  \} \mid \varphi, \{y_i \})$, is the probability of observing the trajectories within the dataset given the candidate specification. It is reasonable to assume that the demonstrations are independent of each other; thus, the total likelihood can be factored as follows:

 $P(\{ \left[ \bm{\alpha} \right]_i  \} \mid \varphi, \{ y_i\}) = \prod_{i\in\{1,2,\hdots,n_{demo}\}} P(\varphi) P([\bm{\alpha}]_i \mid \varphi, y_i)$.

The probability of observing a given trajectory demonstration is dependent upon the underlying dynamics of the domain and the characteristics of the agents producing the demonstrations. In the absence of this knowledge, our aim is to develop an informative, domain-independent proxy for the true likelihood function based only on the properties of the candidate formula; we call this the `complexity-based' (CB) likelihood function. Our approach is founded upon the classical interpretation of probability championed by \cite{laplace1951philosophical}, which involves computing probabilities in terms of a set of equally likely outcomes. Let there be $N_{conj}$ conjunctive clauses in $\varphi$; there are then $2^{N_{conj}}$ possible outcomes in terms of the truth values of the conjunctive clauses. In the absence of any additional information, we assign equal probabilities to each of the potential outcomes. Then, according to the classical interpretation of probability, for candidate formula $\varphi_1$ (defined by subsets $ \bm{\tau_1}, \bm{W_{1_1}}$, and $\bm{W_{2_1}1}$) and $\varphi_2$ (defined by subsets $\bm{\tau_2}, \bm{W_{1_2}}$, and $\bm{W_{2_2}}$) the likelihood odds ratio if $y_i =1$ is defined as follows:

\begin{equation}
  \frac{P([\bm{\alpha}]_i \mid \varphi_1)}{P([\bm{\alpha}] \mid \varphi_2)} =
  \begin{cases}
  \frac{2^{N_{conj_1} } }{2^{N_{conj_2}}} =
  \frac{2^{|\bm{\tau_1}| + |\bm{W_{1_1}}| + |\bm{W_{2_1}}|}}{2^{|\bm{\tau_2}| + |\bm{W_{1_2}}| + |\bm{W_{2_2}}|}} &, [\bm{\alpha}] \models \varphi_2 \\
  \frac{2^{N_{conj_1}}}{\epsilon} =
  \frac{2^{|\bm{\tau_1}| + |\bm{W_{1_1}}| + |\bm{W_{2_1}}|}}{\epsilon} &, [\bm{\alpha}] \nvDash \varphi_2
\end{cases}
  \label{Eq:Custom}
\end{equation}

Here, a finite probability proportional to $\epsilon$ is assigned to a demonstration that does not satisfy the given candidate formula. With this likelihood distribution, a more-restrictive formula with a low prior probability can gain favor over a simpler formula with higher prior probability given a large number of observations that would satisfy it. However, if the candidate formula is not the true specification, a larger set of demonstrations is more likely to include non-satisfying examples, thereby substantially decreasing the posterior probability of the candidate formula. The design of this likelihood function is inspired by the size principle described by \cite{tenenbaum2000rules}.

A second choice for a likelihood function, inspired by \cite{Shepard1317}, is defined as the SIM model by \cite{tenenbaum2000rules}; we call this the `complexity-independent' (CI) likelihood function, and it is defined as follows:
\begin{equation}
  P([\bm{\alpha}] \mid \varphi) =
  \begin{cases}
    1-\epsilon, &\text{if}~ [\bm{\alpha}] \models \varphi \\
    \epsilon, & \text{Otherwise}
  \end{cases}
  \label{Eq:Constant}
\end{equation}

We must define likelihood functions for both acceptable and unacceptable demonstrations. Note that the likelihood function defined by Equation \ref{Eq:Custom} produces a relatively larger likelihood value if the candidate formula correctly classifies the demonstration, and a very small likelihood value if it does not. Following the classical probability argument as before, with $2^{N_{conj}}$ conjunctive clauses in a candidate formula, there are $2^{N_{conj}}$ possible evaluations of each of the individual clauses that would result in the given demonstration not satisfying the candidate formula. Thus, the likelihood function for $y_i=0$ is defined as follows:

\begin{equation}
  \frac{P([\bm{\alpha}]_i \mid \varphi_1)}{P([\bm{\alpha}]_i \mid \varphi_2)} =
  \begin{cases}
  \frac{ 2^{N_{conj_1}}  (2^{N_{conj_2}} - 1) }{ 2^{N_{conj_2}} (2^{N_{conj_1}}-1) }  &, [\bm{\alpha}] \nvDash \varphi_2 \\
  \frac{2^{N_{conj_1}}}{ (2^{N_{conj_1}}-1)\epsilon } &,  [\bm{\alpha}] \models \varphi_2
\end{cases}
  \label{Eq:CustomNeg}
\end{equation}

An equivalent SIM likelihood function for examples with $y_i=0$ is defined as follows:
\begin{equation}
  P([\bm{\alpha}] \mid \varphi) =
  \begin{cases}
    1-\epsilon, &\text{if}~ [\bm{\alpha}] \nvDash \varphi \\
    \epsilon, & \text{Otherwise}
  \end{cases}
  \label{Eq:ConstantNeg}
\end{equation}

Note that for larger values of $N_{conj_1}$ and $N_{conj_2}$ and a negative label $y_i = 1$, the difference between the CI and the CB likelihood function is very small.

\subsubsection{Inference}
\label{SS:Inference}
We implemented our probabilistic model in webppl (\cite{dippl}), a Turing-complete probabilistic programming language. The hyperparameters, including those defined in \autoref{Tab:priors} and $\epsilon$, were set as follows: $p_E,p_G = 0.8$; $p_{part} = 0.3$; $N_{new} = 5$; $\epsilon = 4\times log(2)\times(|\bm{T}+|\bm{\Omega}| +0.5|\bm{\Omega}|(|\bm{\Omega}|-1))$. These values were held constant for all evaluation scenarios. The equation for $\epsilon$ was defined such that evidence of a single non-satisfying demonstration would negate the contribution of four satisfying demonstrations to the posterior probability. The posterior distribution of candidate formulas is constructed using webppl's Markov chain Monte Carlo (MCMC) sampling algorithm from 10,000 samples, with 100 samples serving as burn-in. The posterior distribution is stored as a categorical distribution, with each possibility representing a unique formula. The maximum a posteriori (MAP) candidate represents the best estimate for the specification as per the model. We ran the inference on a desktop with an Intel i7-7700 processor.

\section{Evaluations}
\label{Sec:Eval}

We evaluated the performance of our model across three different domains. We developed a synthetic domain with a low dimensional state-space where we could easily vary the complexity of the ground-truth specifications. We also applied our model to a real-world task of setting a dinner table --  a task often incorporated into studies of learning from demonstrations (\cite{toris2015unsupervised}). This task has a large raw state-space incorporating the poses of the objects included in the domain. This domain demonstrates the benefits of using propositions to represent task specifications, as the complexity of the problem depends upon the number of Boolean propositions and not the dimensionality of the raw state-space. (Note that the ground-truth specifications are known in both of these domains, and it is easy to measure the quality of the solution by comparing it to the ground-truth specification.)

Finally, we also applied our inference model to the large-force exercise (LFE) domain. Large-force exercises are simulated air-combat games used to train combat pilots. We developed simulation environments using joint semi-automated forces (JSAF), a constructive environment for generating examples of LFE executions, and used our model to infer specifications for successful completion of mission objectives. In this domain, the true specifications are not known, and we only have annotations of the demonstrated scenario from a subject matter expert (in this case, the mission commander who designs the scenario and debriefs participating pilots).

%a domain representing a real-world task of setting a dinner table -- a task often incorporated into studies of learning from demonstrations (\cite(toris2015unsupervised)), and the large force exercise domain

%We evaluated the performance of our model within two different domains: a synthetic domain in which we could easily vary the complexity of the ground truth specifications, and a domain representing the real-world task of setting a dinner table — a task often incorporated into studies of learning from demonstration (\cite{toris2015unsupervised}).

\subsection{Metrics}

The evaluation metrics used to test the quality of the inferred specifications depend upon whether the ground-truth specifications are known. For domains in which it is known (the synthetic and dinner-table domains), the ground-truth specification is defined using subsets $\bm{\tau^*}, \bm{W_1^*}$, and $\bm{W_2^*}$ (as per Equations \ref{Eq:Global}, \ref{Eq:Eventually}, and \ref{Eq:Order}), and a candidate formula $\varphi$ is defined by subsets $\bm{\tau}, \bm{W_1}$, and $\bm{W_2}$. In such cases, we define the degree of similarity using the Jaccard index (\cite{jaccard}) as follows:

\begin{equation}
  L(\varphi) = \frac{\mid \{\bm{\tau^*} \cup \bm{W_1^*} \cup \bm{W_2^*}\} \cap \{ \bm{\tau} \cup \bm{W_1} \cup \bm{W_2} \} \mid}{\mid \{\bm{\tau^*} \cup \bm{W_1^*} \cup \bm{W_2^*} \} \cup \{ \bm{\tau} \cup \bm{W_1} \cup \bm{W_2} \} \mid}
  \label{Eq:Jaccard}
\end{equation}

The maximum possible value of $L(\varphi)$ is one such that both formulas are equivalent. One key benefit of our approach is that we compute a posterior distribution over candidate formulas; thus, we report the expected value of $\mathbb{E}(L(\varphi))$ as a measure of the deviation of the inferred distribution from the ground truth. We also report the maximum value of $L(\varphi)$ among the top 5 candidates in the posterior distribution. We classify the inferred orders in $\bm{W_2}$ as correct if they are included in the ground truth, incorrect if they reverse any constraint within the ground truth, and `extra' otherwise. (Extra orders over-constrain the problem, but do not induce incorrect behaviors.)

For the LFE domain, where the ground-truth specifications are unknown but SME annotations for whether the mission objectives were accomplished are provided for the dataset, we use the weighted F1 score for both `achieved' and `failed' labels. This score is evaluated on a test set that is held out while using the remaining examples in the dataset to infer the specifications.

%We evaluated our approach against the temporal logic inference (TempLogIn) algorithm proposed by \cite{kong2017temporal}. TempLogIn mines parametric STL (PSTL) specifications, by conducting a breadth first search through a DAG induced by a partial ordering relation between PSTL formulas. Note that while our approach requires only positive examples, temporal logic inference must be trained on both positive and negative examples.

\begin{figure}
    \centering
    \begin{subfigure}[b]{0.24\textwidth}
        \centering
        \includegraphics[width=\textwidth]{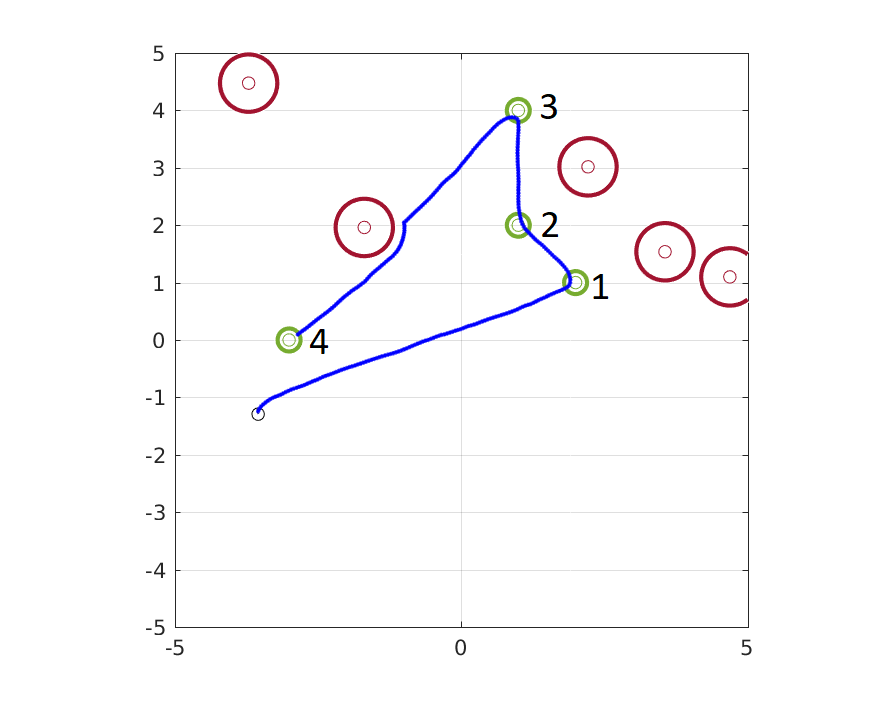}
        %\caption{}

    \end{subfigure}
    %\hfill
    \begin{subfigure}[b]{0.24\textwidth}
        \centering
        \includegraphics[width=\textwidth]{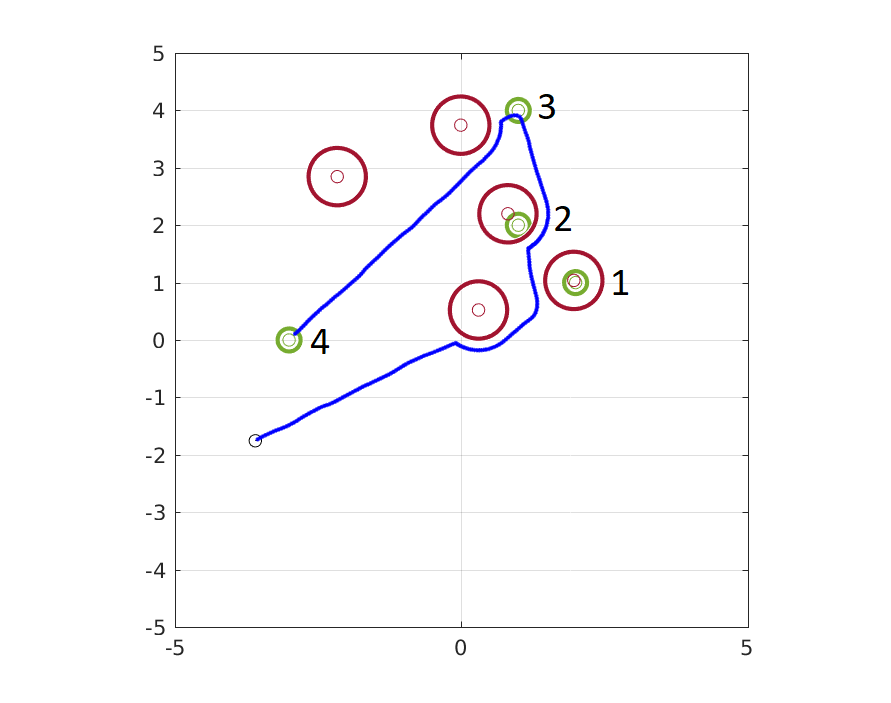}
        %\caption{}

    \end{subfigure}
    %\vskip\baselineskip
    \begin{subfigure}[b]{0.24\textwidth}
        \centering
        \includegraphics[width=\textwidth]{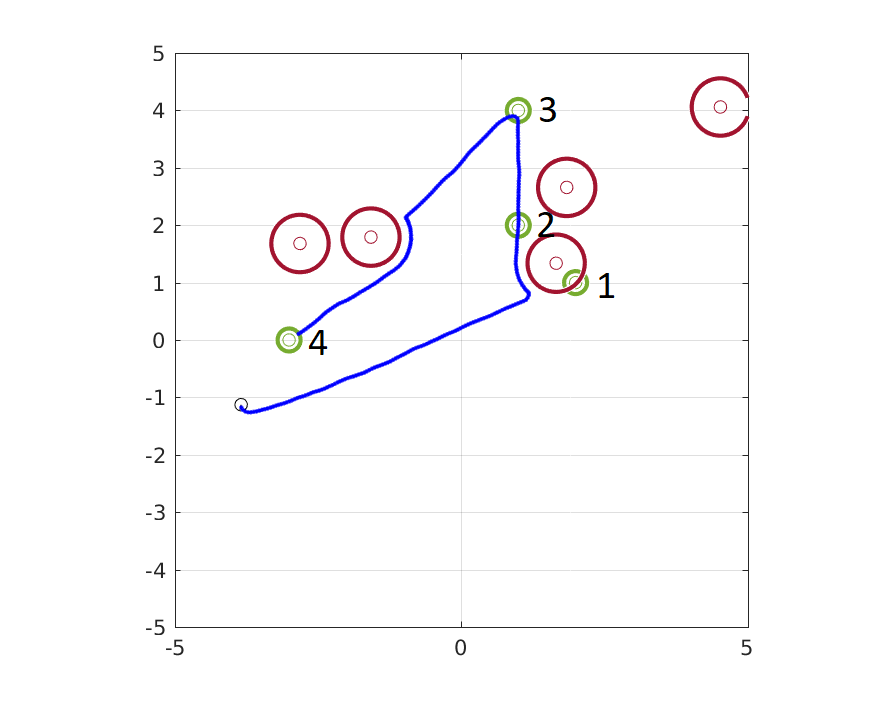}
        %\caption{}%

    \end{subfigure}
    %\quad
    \begin{subfigure}[b]{0.24\textwidth}
        \centering
        \includegraphics[width=\textwidth]{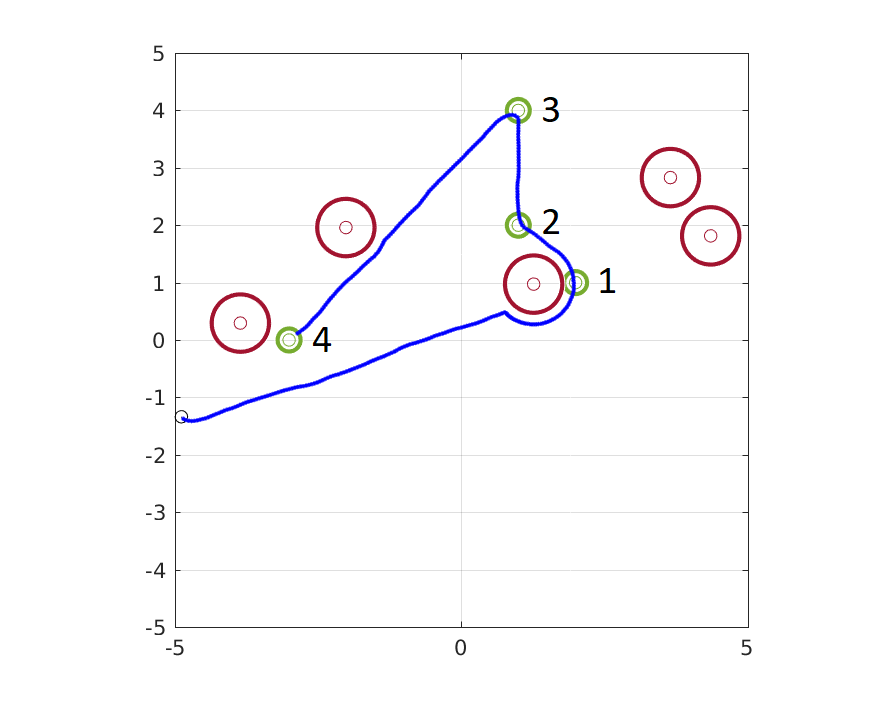}
%
        %\caption{}%

    \end{subfigure}
    \caption{Example trajectories from Scenario 1. Green circles denote the POIs; red circles denote the avoidance zones of threats.}
    \label{fig:D2}
    %\vspace{-15pt}
\end{figure}

\subsection{Synthetic Domain}

In our synthetic domain, an agent navigates within a two-dimensional space that includes points of interest (POIs) to visit and threats to avoid. The state of the agent $\bm{x}$ represents the position of that agent within the task space.

Let $\bm{\tau} = \{1,2 \hdots, n_{threats} \}$ represent a set of threats positioned at $\bm{x}_{T_i} ~\forall ~i \in \bm{\tau}$, respectively. A proposition $\tau_i$ is associated with each threat location $i\in \bm{tau}$ such that:

\newcommand\norm[1]{\left\lVert#1\right\rVert}

\begin{equation}
  \tau_i = \begin{cases}
            \text{true}, & \norm{\bm{x} - \bm{x}_{T_i}} \geq \epsilon_{threat} \\
            \text{false}, & \text{otherwise}
           \end{cases}
\end{equation}

The proposition $\tau_{T_i}$ holds if the agent is not within the avoidance radius $\epsilon_{threat}$ of the threat location.

Let $\bm{\Omega} = \{1,2,\hdots, n_{POI}\}$ represent the set of POIs positioned at $\bm{x}_{P_i} ~\forall~i\in\bm{\Omega}$. A proposition $\omega_i$ is associated with each POI such that:

\begin{equation}
  \omega_i = \begin{cases}
                \text{true}, & \norm{\bm{x} - \bm{x}_{P_i}} \leq \epsilon_{POI} \\
                \text{false}, & \text{otherwise}
             \end{cases}
\end{equation}

$\omega_i$ evaluates as true if the agent is within a tolerance radius $\epsilon_{POI}$ of the POI.

Finally, propositions $\pi_i ~\forall~ i \in \bm{\Omega}$ are conditions propositions that denote the accessibility of the POI $i$, and are defined as follows:

\begin{equation}
  \pi_i = \begin{cases}
            \text{false}, & \exists ~j ~\text{such that} \norm{ \bm{x}_{P_i} - \bm{x}_{T_j} }\leq \epsilon_{threat}\\
            \text{true}, & \text{otherwise}
          \end{cases}
\end{equation}

$\pi_i$ evaluates as false if the POI $i$ is inside the avoidance region of any of the threats.

The agent can be programmed to visit the accessible POIs and avoid threats as per the ground-truth specification. The ground-truth specifications are stated by defining the following: a set $\bm{T} \subseteq \bm{\tau}$ that represents the subset of threats that the agent must avoid; a set $\bm{W_1}\subseteq \bm{\Omega}$ that represents the subset of POIs the agent must visit; and the ordering constraints defined by $\bm{W_2}$, a set of feasible pairwise precedence constraints between the POIs.

Here, we demonstrate the results of applying our inference model to three scenarios with differing ground-truth specifications.

\textbf{Scenario 1}: In Scenario 1, we placed five threats in the task-domain, and their positions were sampled from a uniform distribution for each demonstration. There were four points of interest, labeled ${1,2,3,4}$, and their positions were fixed across all demonstrations. The agents were required to visit the POIs in a fixed order ($[1,2,3,4]$). Example trajectories from this scenario are depicted in \autoref{fig:D2}.

\begin{figure}
     \centering
     \begin{subfigure}[b]{0.3\textwidth}
         \centering
         \includegraphics[width=\textwidth]{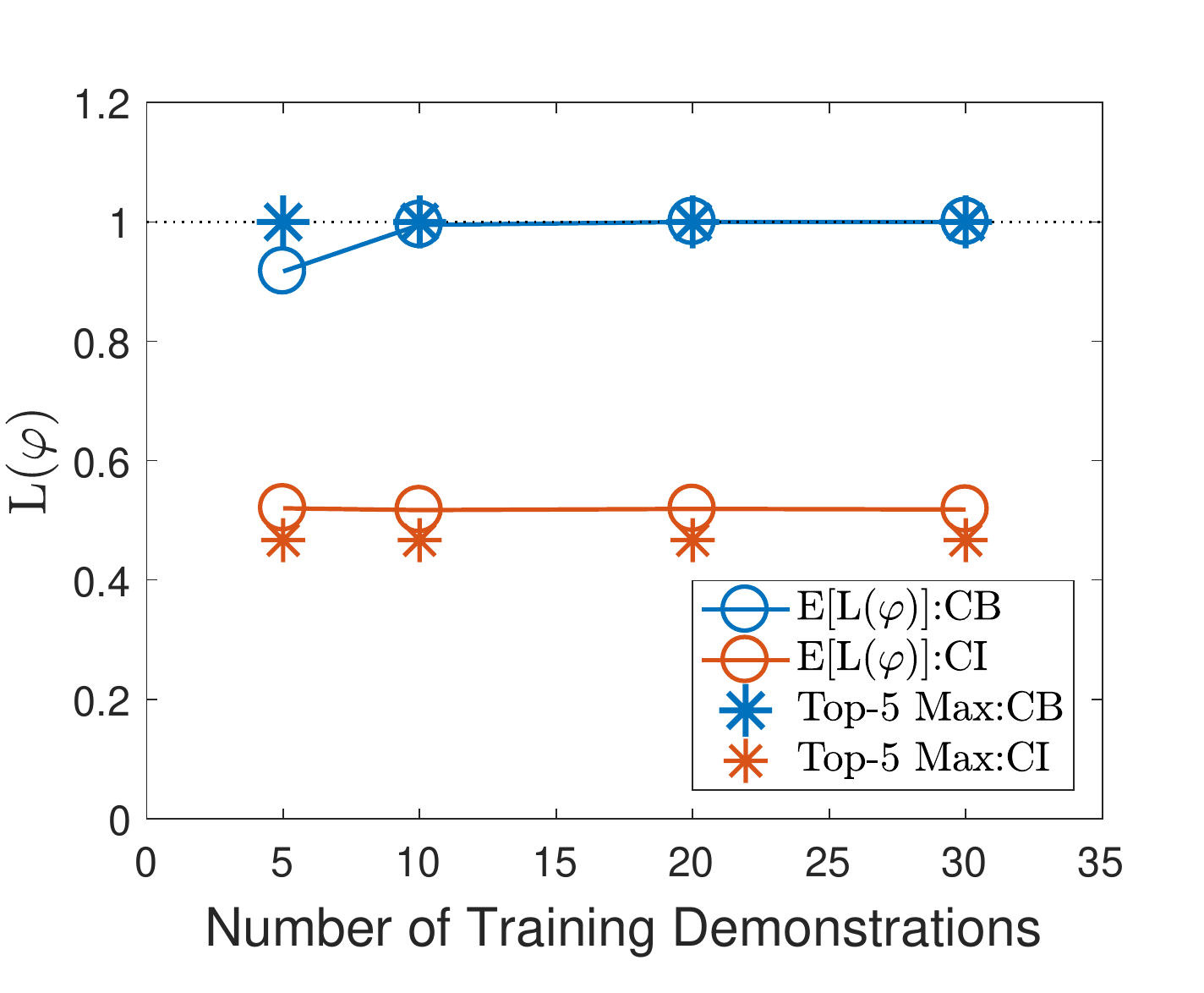}
         \caption{Scenario 1}
         \label{fig:Res1A}
     \end{subfigure}
     %\hfill
     \begin{subfigure}[b]{0.3\textwidth}
         \centering
         \includegraphics[width=\textwidth]{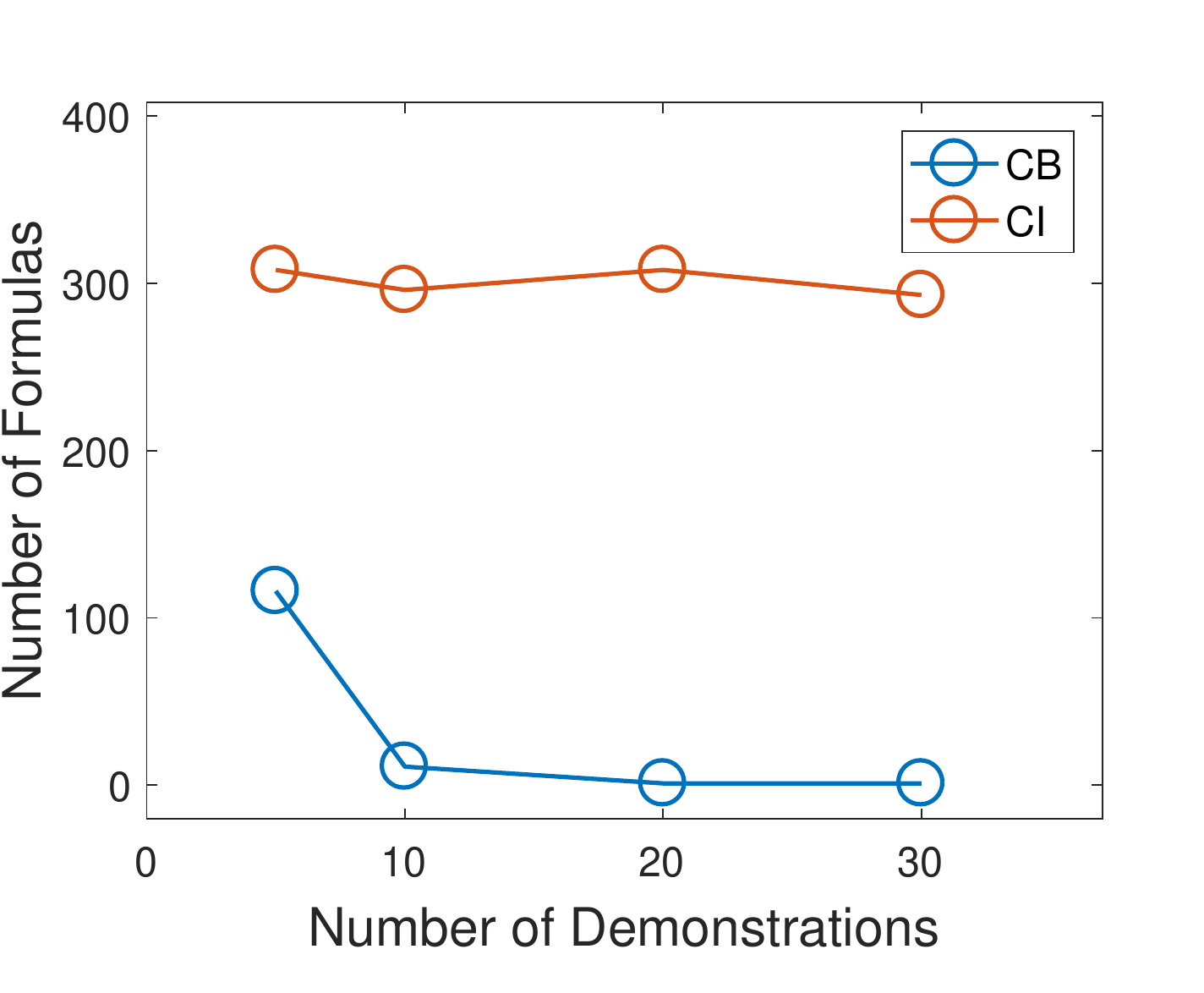}
         \caption{Scenario 1}
         \label{fig:Res1B}
     \end{subfigure}
     \caption{\autoref{fig:Res1A} depicts the results from Scenario 1, with the dotted line representing the maximum possible value of $L(\varphi)$. \autoref{fig:Res1B} shows the number of unique formulas in the posterior distribution}
     \label{fig:res1}
\end{figure}

The posterior distribution was computed using prior 1 (defined in \autoref{Tab:priors}), with both CB (\autoref{Eq:Custom}) and CI (\autoref{Eq:Constant}) likelihood functions. The expected and maximum values among the top 5 a posteriori formula candidates of $L(\varphi)$ are depicted in \autoref{fig:res1}. We observed that the CB likelihood function performed better than the CI likelihood function at inferring the complete specification. Using the CI function resulted in a higher posterior probability assigned to formulas with high prior probability that were satisfied by all demonstrations. (These tended to be simple, non-informative formulas; the CB function assigned higher probability mass to more-complex formulas that explained the demonstrations correctly.) \autoref{fig:Res1B} depicts the number of unique formulas in the posterior distributions. The CB likelihood function resulted in posteriors being more peaky, with fewer unique formulas as training set size increased; this effect was not observed with the CI function.

\begin{figure}
     \centering
     \begin{subfigure}[b]{0.3\textwidth}
         \centering
         \includegraphics[width=\textwidth]{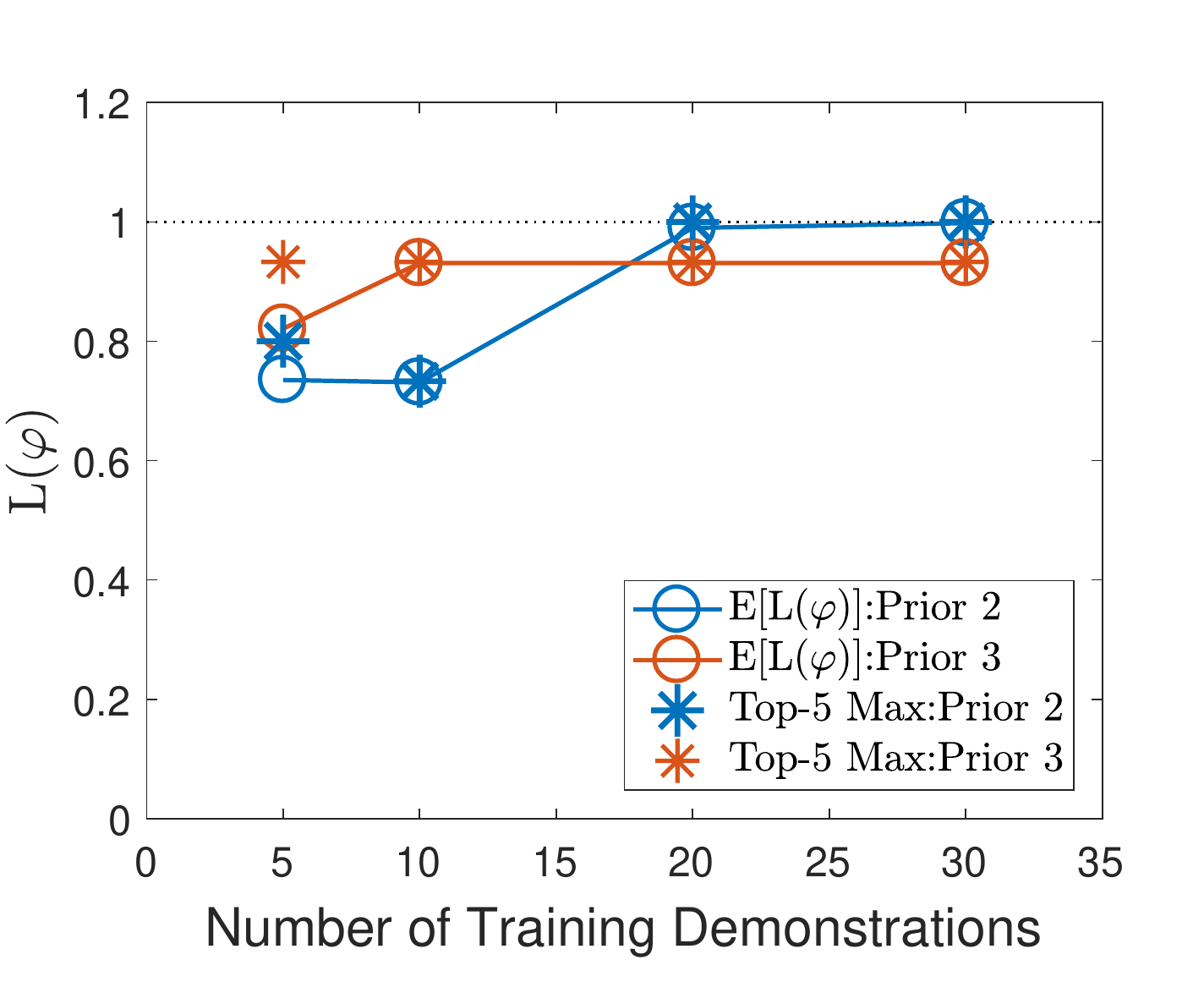}
         \caption{Scenario 1}
         \label{fig:Res1p23A}
     \end{subfigure}
     %\hfill
     \begin{subfigure}[b]{0.3\textwidth}
         \centering
         \includegraphics[width=\textwidth]{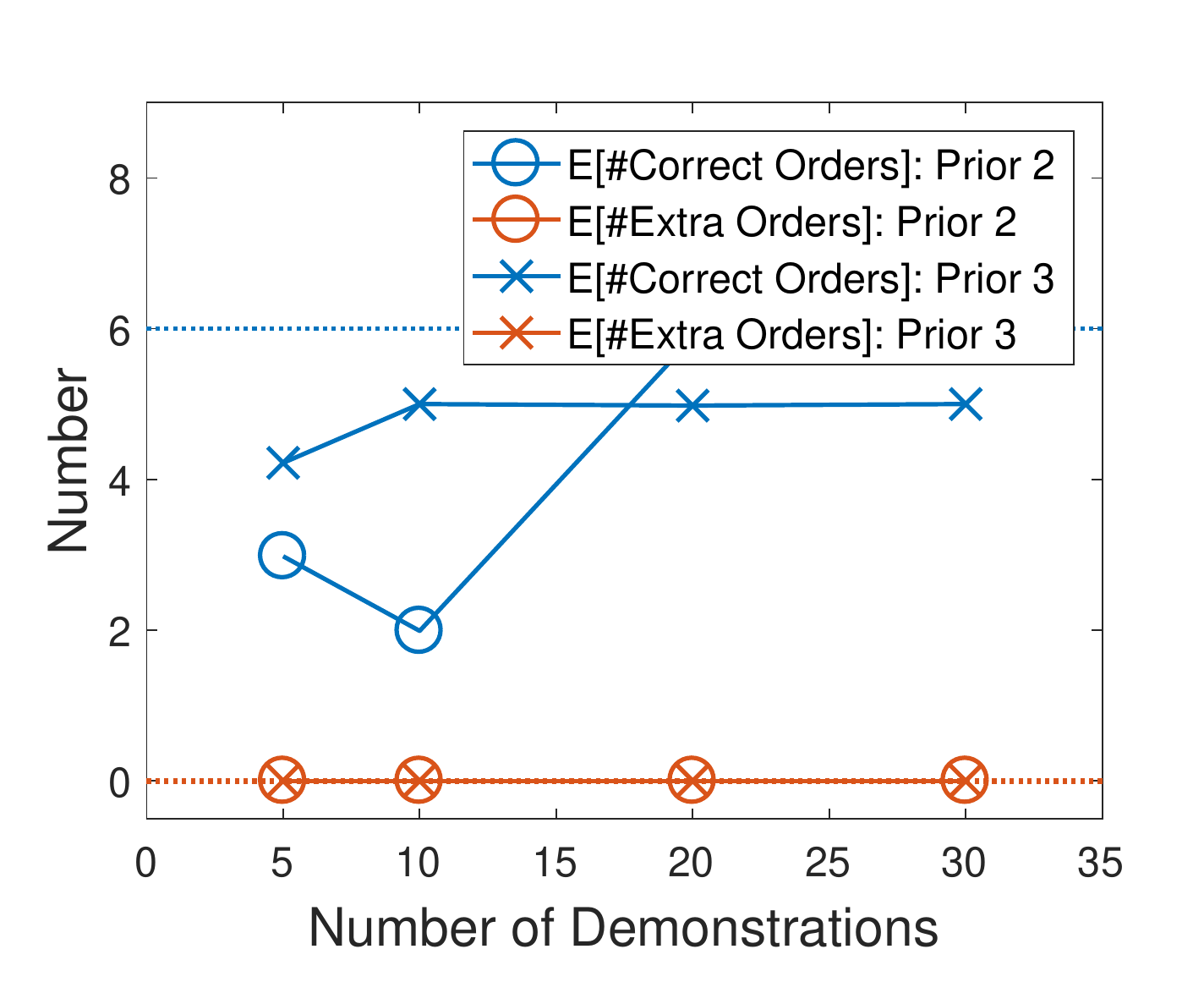}
         \caption{Scenario 1}
         \label{fig:Res1p23B}
     \end{subfigure}
     \caption{\autoref{fig:Res1p23A} depicts the results from Scenario 1 using priors 2 and 3, with the dotted line representing the maximum possible value of $L(\varphi)$. \autoref{fig:Res1p23B} depicts the expected value of the number of correct and extra orders in the posterior distribution.}
     \label{fig:res1p23}
\end{figure}

The posterior distribution was also computed using priors 2 and 3 with the CB likelihood function. The expected and maximum values among the top 5 a posteriori formula candidates of $L(\varphi)$ are depicted in \autoref{fig:Res1p23A}. Prior 3 aligned better with the ground-truth specification with fewer training examples. With a larger training set, prior 2 recovered the exact specification, while prior 3 failed to do so. \autoref{fig:Res1p23B} depicts the expected value of the correct and extra orders in the candidate formulas included in the posterior distribution. The a priori bias of prior 3 toward longer chains is apparent, as it recovered more correct orders with fewer training demonstration in comparison to prior 2. Prior 2 recovered all correct priors with more training examples; however, prior 3 failed to do so with 30 training examples.

\begin{figure}
    \centering
    \begin{subfigure}[b]{0.24\textwidth}
        \centering
        \includegraphics[width=\textwidth]{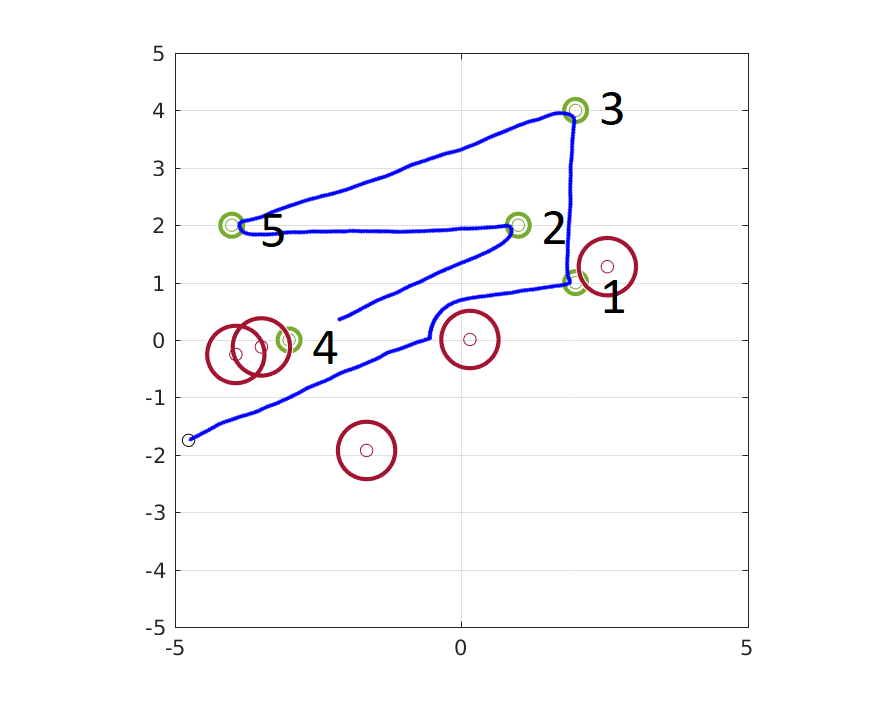}
        %\caption{}

    \end{subfigure}
    %\hfill
    \begin{subfigure}[b]{0.24\textwidth}
        \centering
        \includegraphics[width=\textwidth]{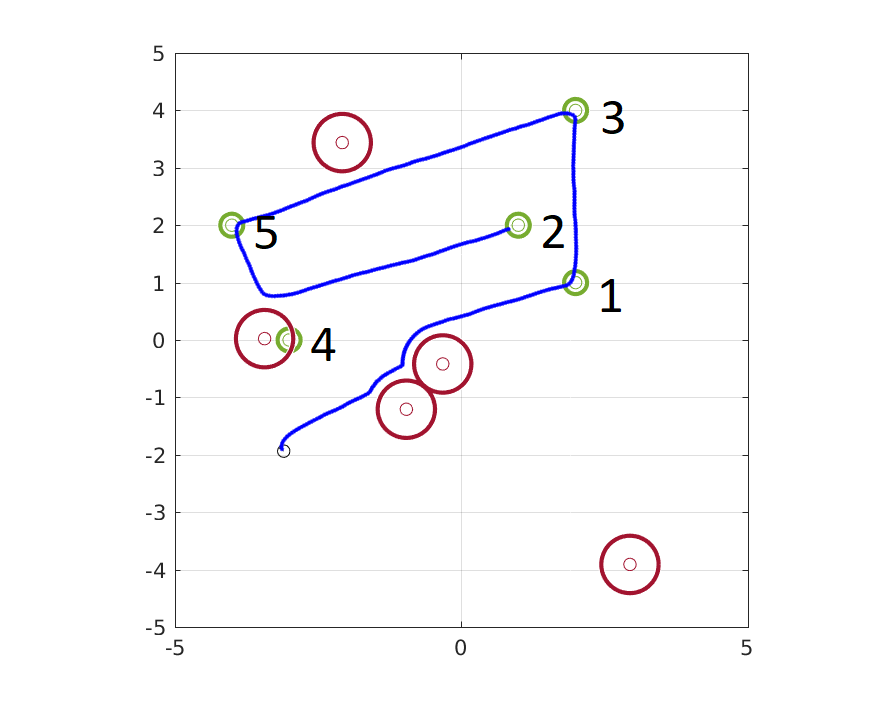}
        %\caption{}

    \end{subfigure}
    %\vskip\baselineskip
    \begin{subfigure}[b]{0.24\textwidth}
        \centering
        \includegraphics[width=\textwidth]{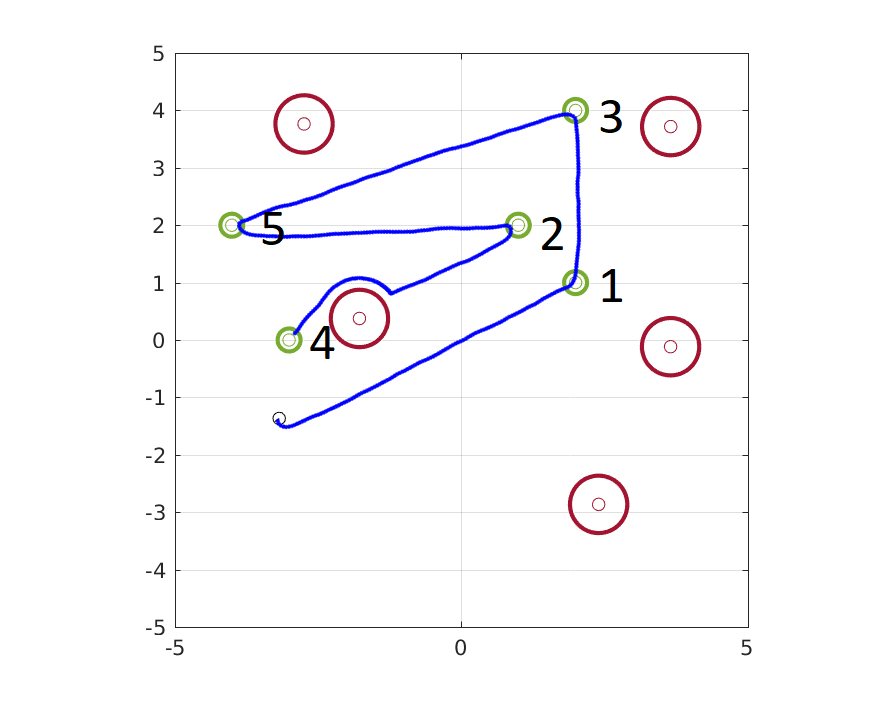}
        %\caption{}%

    \end{subfigure}
    %\quad
    \begin{subfigure}[b]{0.24\textwidth}
        \centering
        \includegraphics[width=\textwidth]{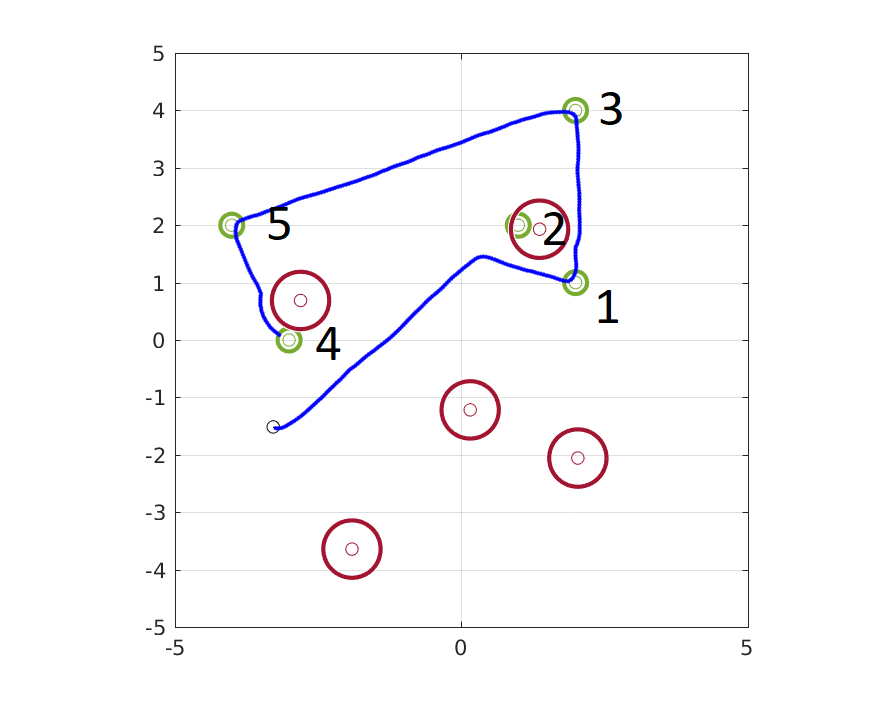}
%
        %\caption{}%

    \end{subfigure}
    \caption{Example trajectories from Scenario 2. Green circles denote the POIs; red circles denote the avoidance zones of threats.}
    \label{fig:scen2}
    %\vspace{-15pt}
\end{figure}

\textbf{Scenario 2}: Scenario 2 contained five POIs ${1,2,3,4,5}$ and five threats. Like Scenario 1, the threat positions were sampled uniformly for each demonstration. All the POIs, if accessible, had to be visited. A partial ordering constraint was imposed such that POIs $[1,3,5]$ had to be visited in that specific order, while POIs $\{ 2, 4\}$ could be visited in any order. Some demonstrations generated for Scenario 2 are depicted in \autoref{fig:scen2}.

\begin{figure}
     \centering
     \begin{subfigure}[b]{0.3\textwidth}
         \centering
         \includegraphics[width=\textwidth]{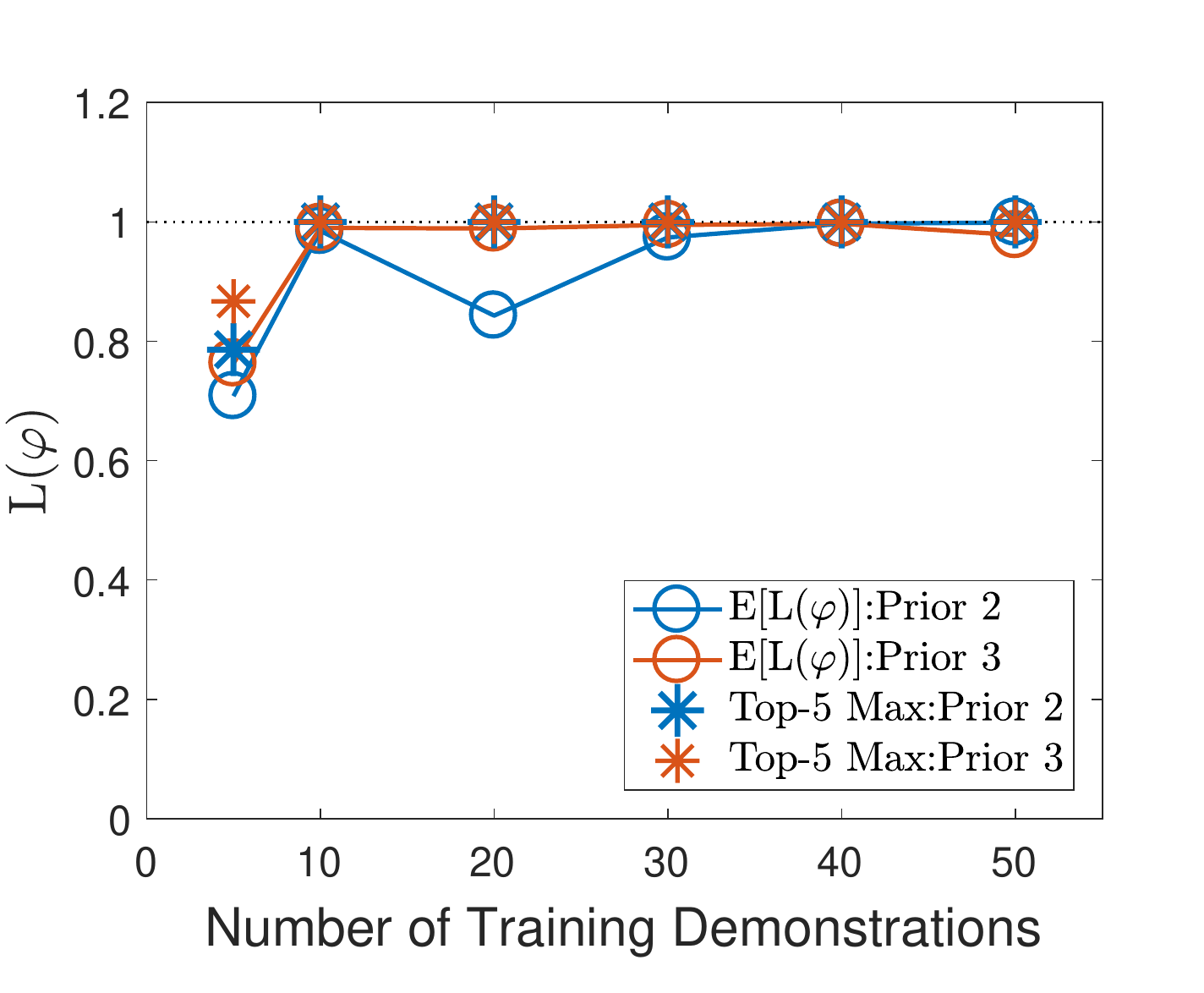}
         \caption{Scenario 2 $L(\varphi)$}
         \label{fig:Res1C}
     \end{subfigure}
     %\hfill
     \begin{subfigure}[b]{0.3\textwidth}
         \centering
         \includegraphics[width=\textwidth]{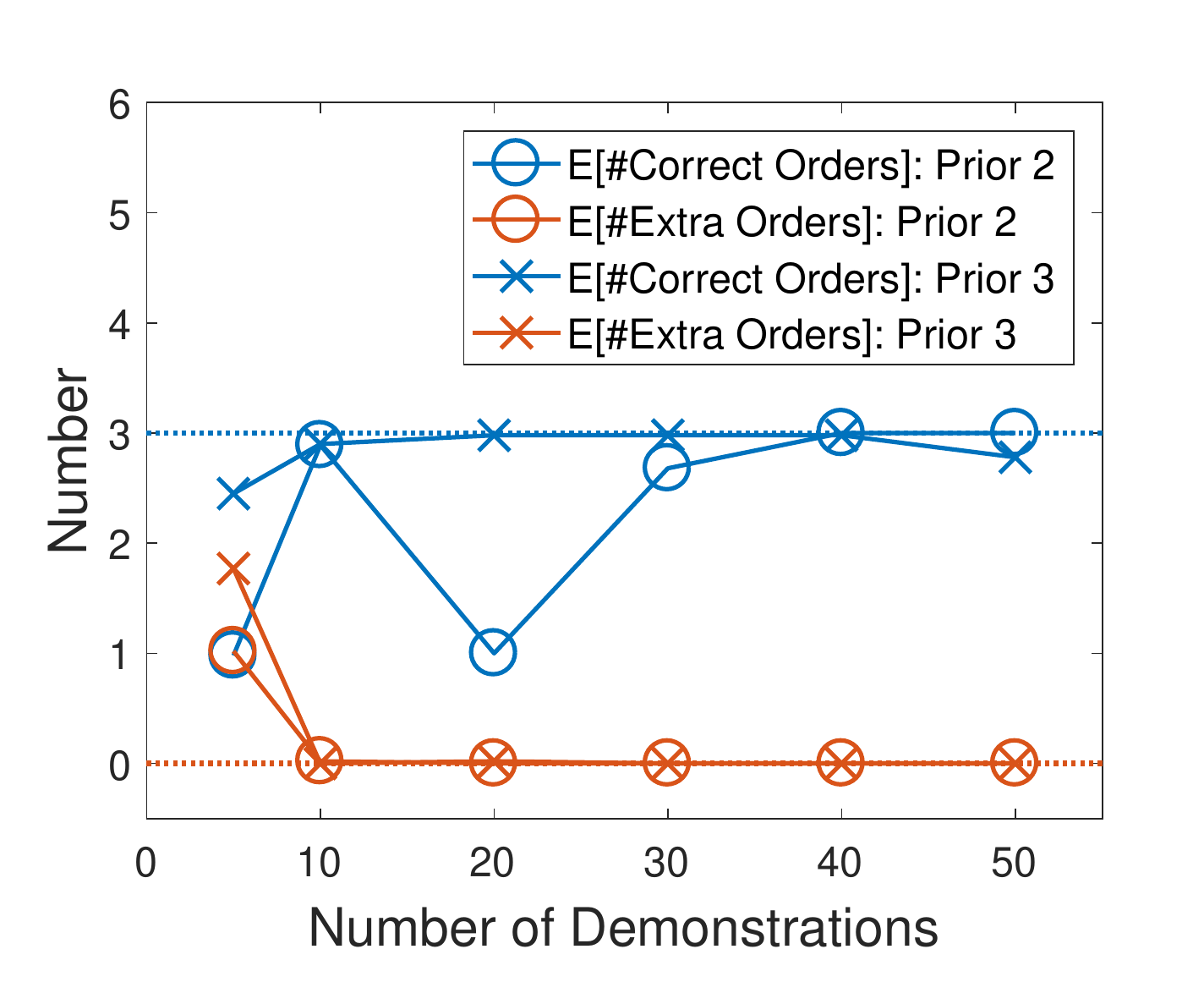}
         \caption{Scenario 2 orders}
         \label{fig:Res1D}
     \end{subfigure}
     %\vskip\baselineskip

     \caption{\autoref{fig:Res1C} indicates the $L(\varphi)$ values for Scenario 2, and \autoref{fig:Res1D} depicts the correct and extra orderings inferred in Scenario 2. The dotted lines represent the number of orderings in the true specification.}
     \label{fig:res2}
\end{figure}

For Scenario 2, the posterior distribution was computed using priors 2 and 3, as the ground-truth specification did not lie in support of prior 1. The expected and maximum values among the top 5 formula candidates of $L(\varphi)$ are depicted in \autoref{fig:Res1C}. Given a sufficient number of training examples, both priors were able to infer the complete formula; with 10 or more training examples, both priors returned the ground-truth formula among the top 5 candidates with regard to posterior probabilities. \autoref{fig:Res1D} depicts the correct and extra orders inferred in Scenario 2. Prior 3 assigned a larger prior probability to longer task chains compared with prior 2, but both priors converged to the correct specification given enough training examples.

\begin{figure}
    \centering
    \begin{subfigure}[b]{0.24\textwidth}
        \centering
        \includegraphics[width=\textwidth]{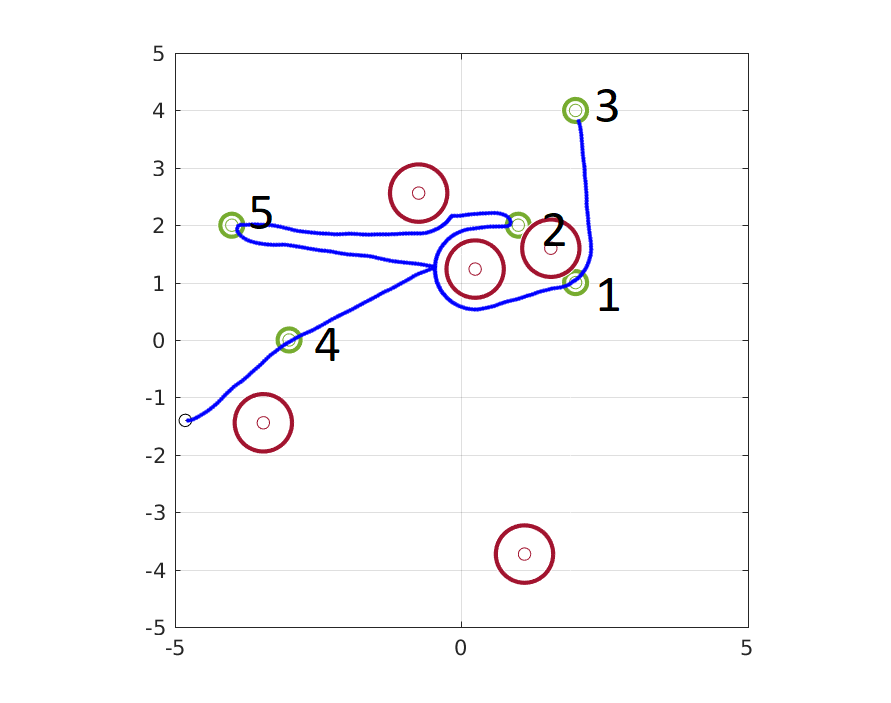}
        %\caption{}
    \end{subfigure}
    %\hfill
    \begin{subfigure}[b]{0.24\textwidth}
        \centering
        \includegraphics[width=\textwidth]{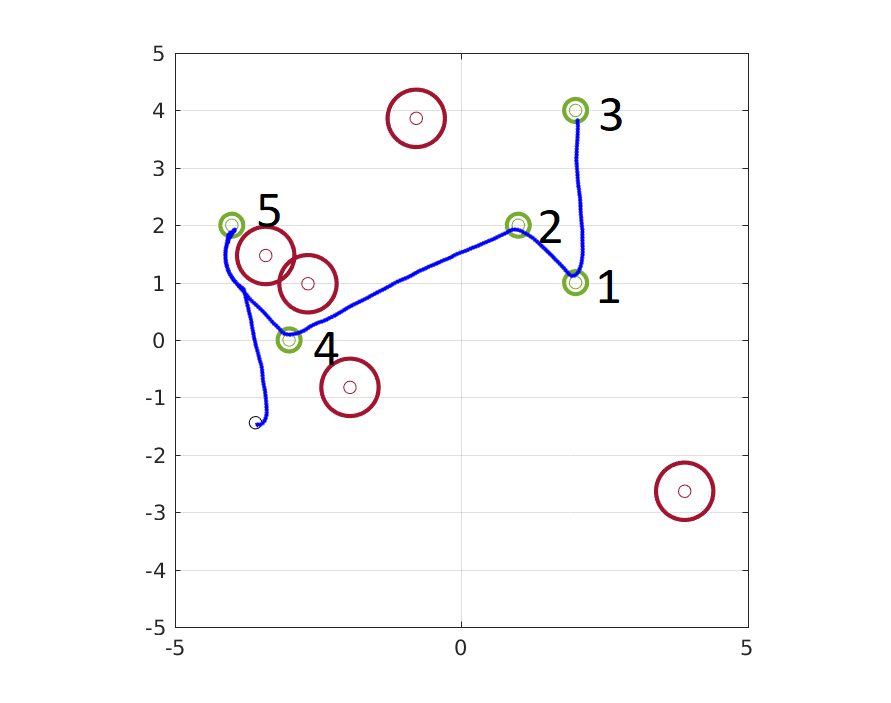}
        %\caption{}
    \end{subfigure}
    %\vskip\baselineskip
    \begin{subfigure}[b]{0.24\textwidth}
        \centering
        \includegraphics[width=\textwidth]{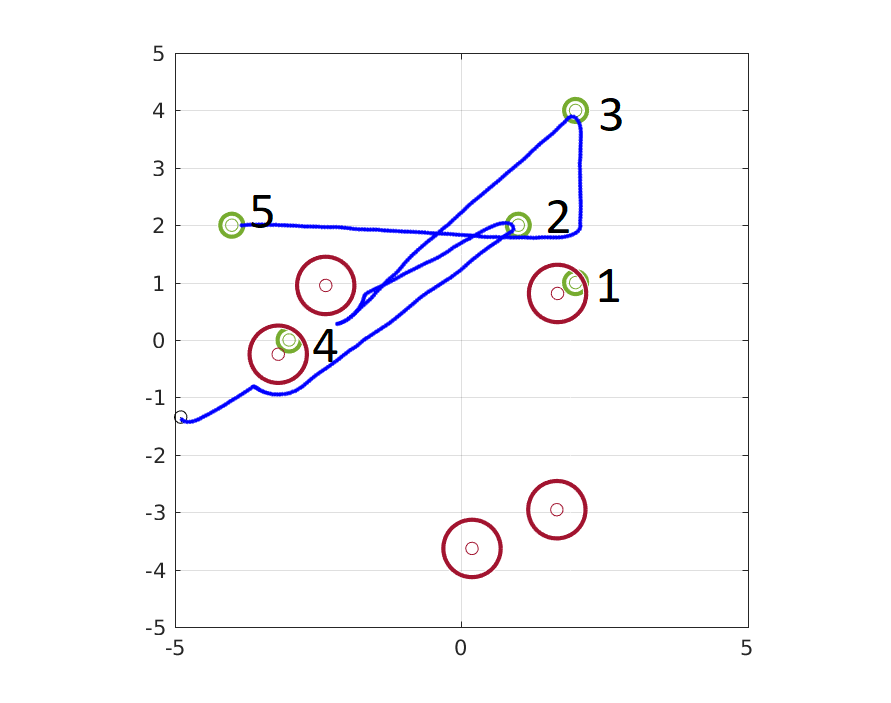}
        %\caption{}%
    \end{subfigure}
    %\quad
    \begin{subfigure}[b]{0.24\textwidth}
        \centering
        \includegraphics[width=\textwidth]{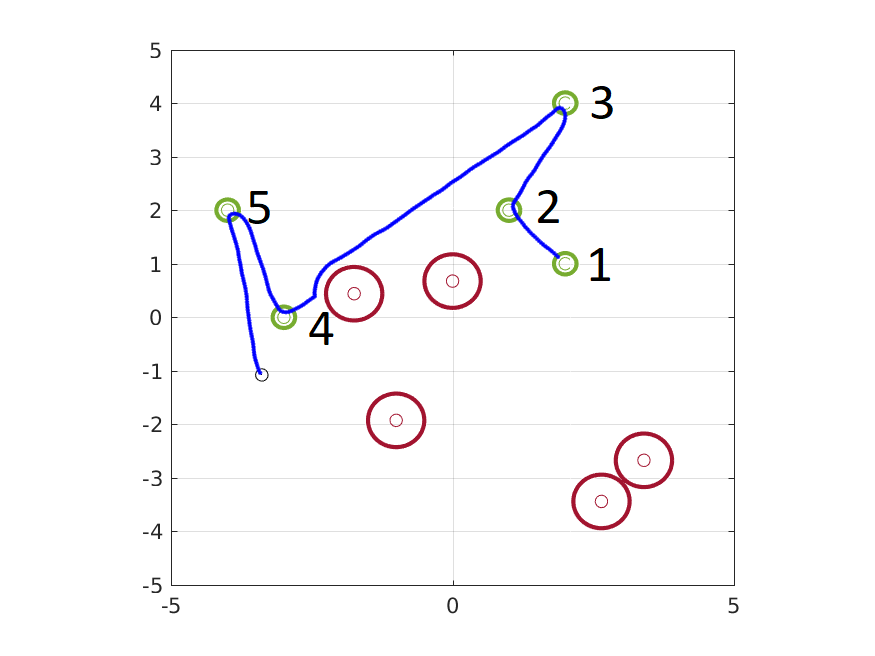}
        %\caption{}%
    \end{subfigure}
    \caption{Example trajectories from Scenario 3. The green circles denote the POIs; the red circles denote the threat avoidance zones.}
    \label{fig:Scen3}
\end{figure}

\textbf{Scenario 3:} Scenario 3 included five threats and five POIs labeled $\{1,2,3,4,5\}$, respectively. The threat positions were uniformly sampled for each scenario. Each of the POIs, if accessible, had to be visited; however, there were no constraints placed on the order in which they were visited. \autoref{fig:Scen3} depicts some of the example demonstrations.

\begin{figure}
     \centering
     \begin{subfigure}[b]{0.3\textwidth}
         \centering
         \includegraphics[width=\textwidth]{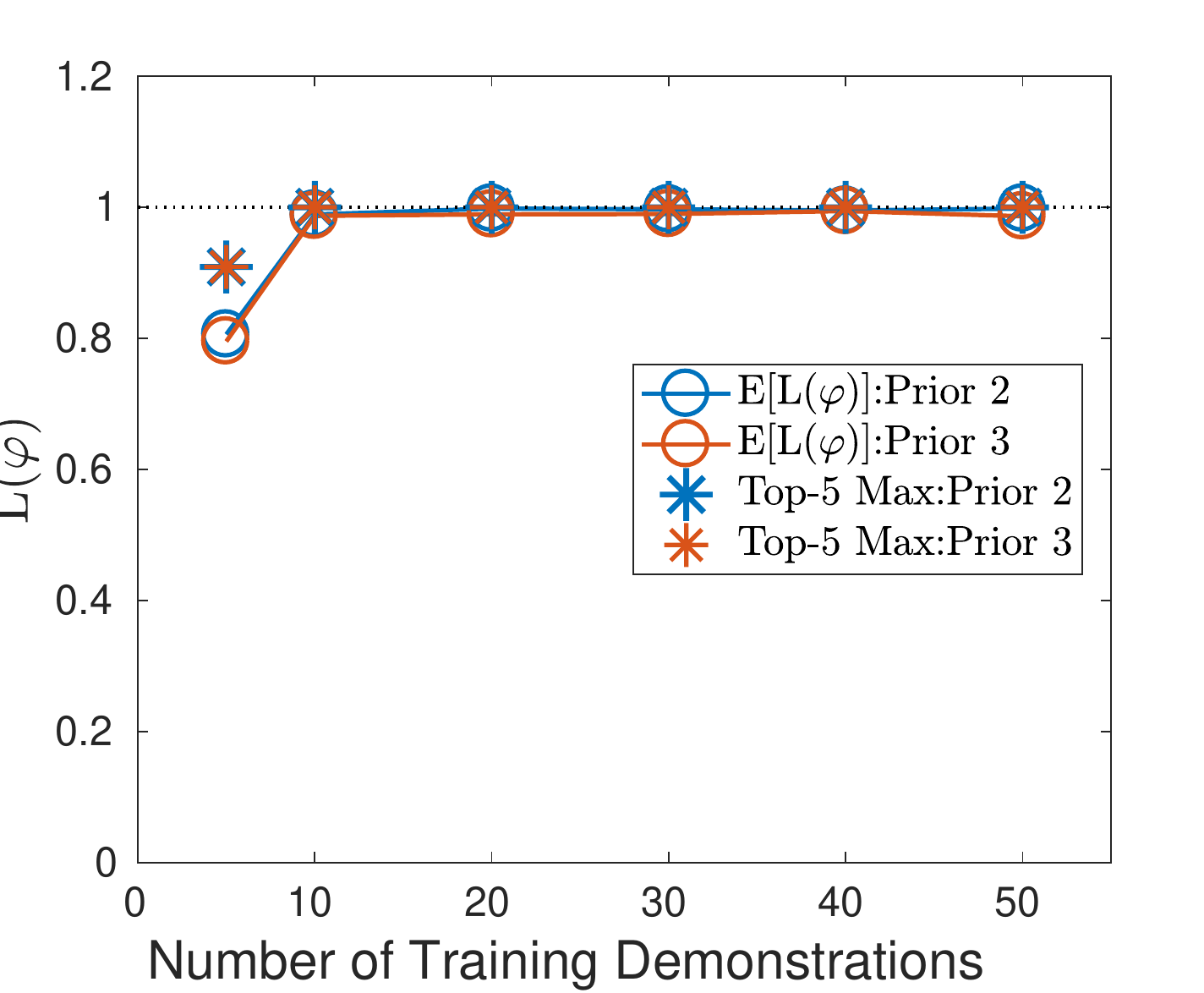}
         \caption{Scenario 2 $L(\varphi)$}
         \label{fig:Res3A}
     \end{subfigure}
     %\hfill
     \begin{subfigure}[b]{0.3\textwidth}
         \centering
         \includegraphics[width=\textwidth]{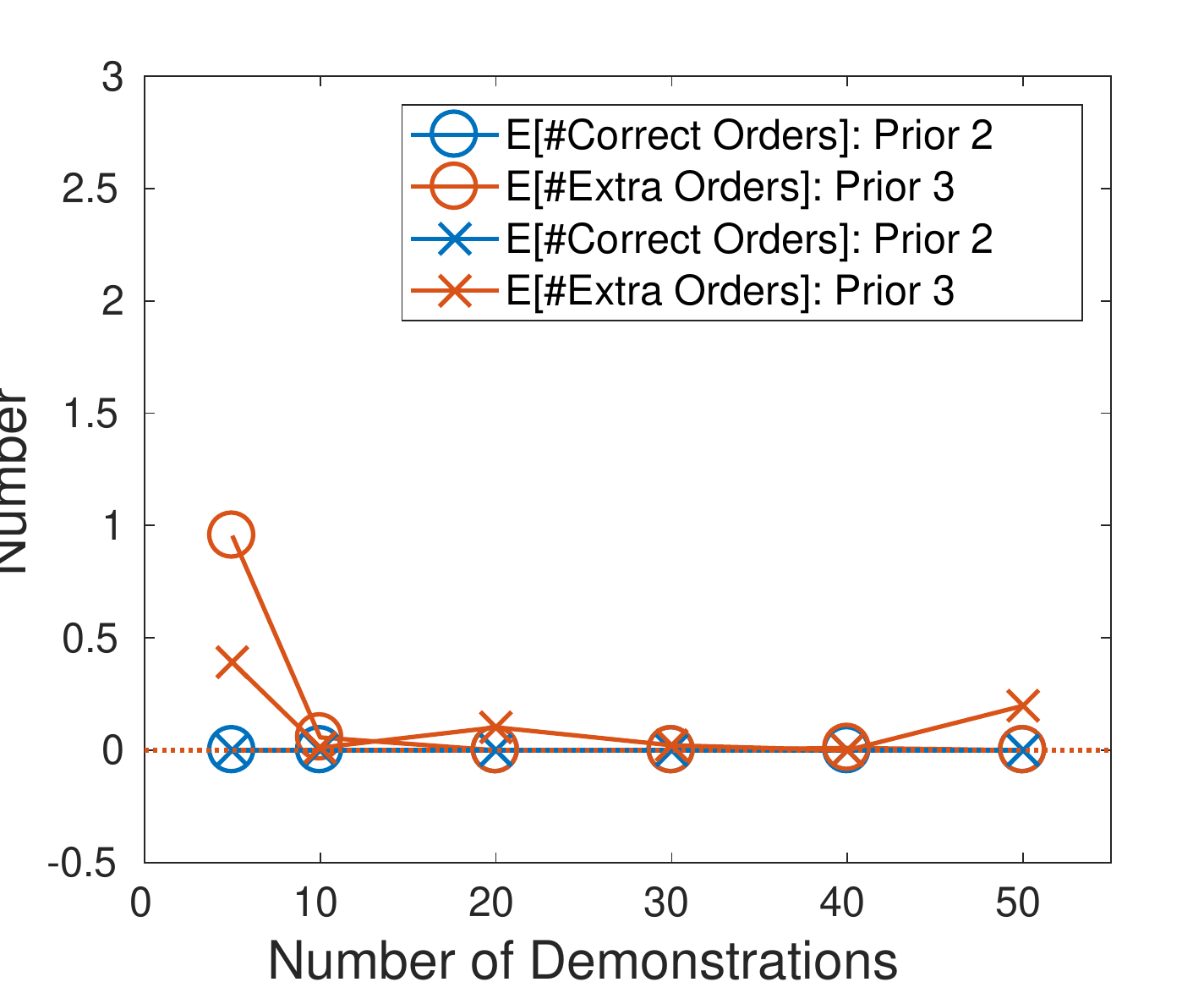}
         \caption{Scenario 2 orders}
         \label{fig:Res3B}
     \end{subfigure}
     %\vskip\baselineskip

     \caption{\autoref{fig:Res3A} indicates the $L(\varphi)$ values for Scenario 3, and \autoref{fig:Res3B} depicts the correct and extra orderings inferred in Scenario 3. The dotted lines represent the number of orderings in the true specification.}
     \label{fig:res3}
\end{figure}

Again, the posterior distribution was computed using priors 2 and 3. The expected and maximum values among the top 5 formula candidates of $L(\varphi)$ are depicted in \autoref{fig:Res3A}. In this scenario, both priors performed equally well with regard to recovering the ground-truth specification. With 10 or more demonstrations, both priors returned the ground-truth specification as the maximum a posteriori estimate. The expected value of the extra orders contained in the posterior distributions is depicted in \autoref{fig:Res3B}. Once again, the tendency of prior 3 to return longer chains is apparent, as more formulas in the posterior distribution returned a greater number of extra ordering constraints as compared with prior 2.

The runtime for MCMC inference is a function of the number of samples generated, the number of demonstrations in the training set, and demonstration length. Scenarios 1 and 2 required an average runtime of 10 and 90 minutes for training set sizes of 5 and 50, respectively.

TempLogIn (\cite{kong2017temporal}) required 33 minutes to terminate with three PSTL clauses. For all the scenarios, the mined formulas did not capture any of the temporal behaviors in Section \ref{SS:FormulaTemplate}, indicating that additional PSTL clauses were required. However, with five and 10 PSTL clauses, the algorithm did not terminate within the 24-hour runtime cutoff. Scaling TempLogIn to larger formula lengths is difficult, as the size of the search graph increases exponentially with the number of PSTL clauses, and the algorithm must evaluate all formula candidates of length $n$ before candidates of length $n+1$.

%\textcolor{blue}{Temporal logic inference returned vacuous specifications when mining formulas with three and five PSTL clauses. It did not terminate after running for 24 hours while mining formulas with 10 PSTL clauses. This was to be expected due to their approach of a breadth-first search over a DAG of candidate formulas along with simulated annealinThe dotted lines represent the true values.g in order to optimize time-windows for each PSTL clause.}

\subsection{Dinner Table Domain}
\begin{figure*}
    \centering
    \begin{subfigure}[b]{0.3\textwidth}
        \centering
        \includegraphics[width=\textwidth]{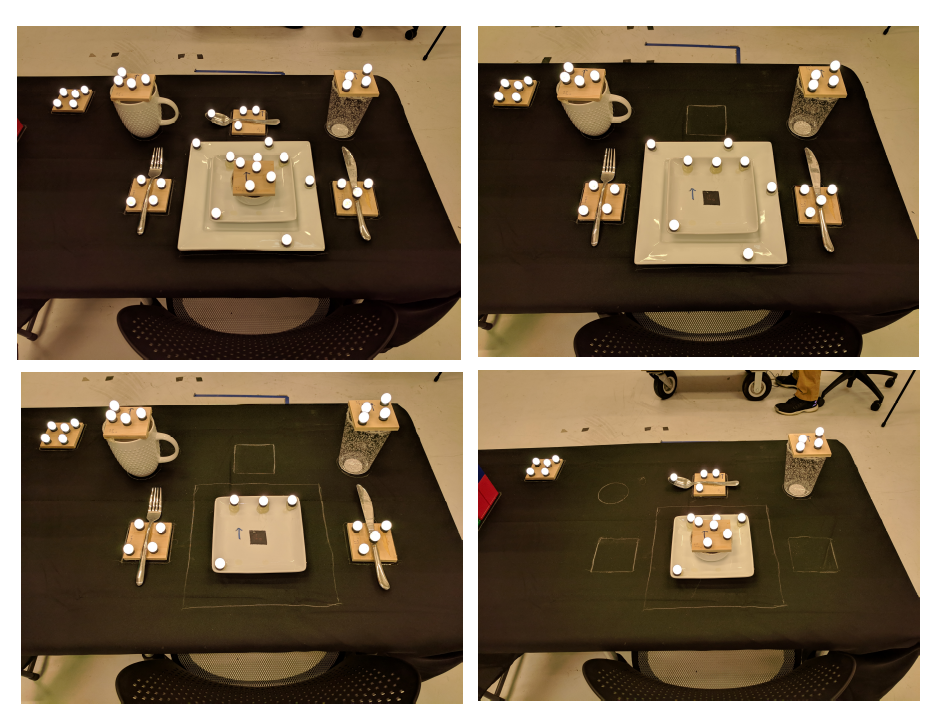}
        \caption{}
        \label{fig:DinnerA}
    \end{subfigure}
    %\hfill
    \begin{subfigure}[b]{0.37\textwidth}
        \centering
        \includegraphics[width=\textwidth]{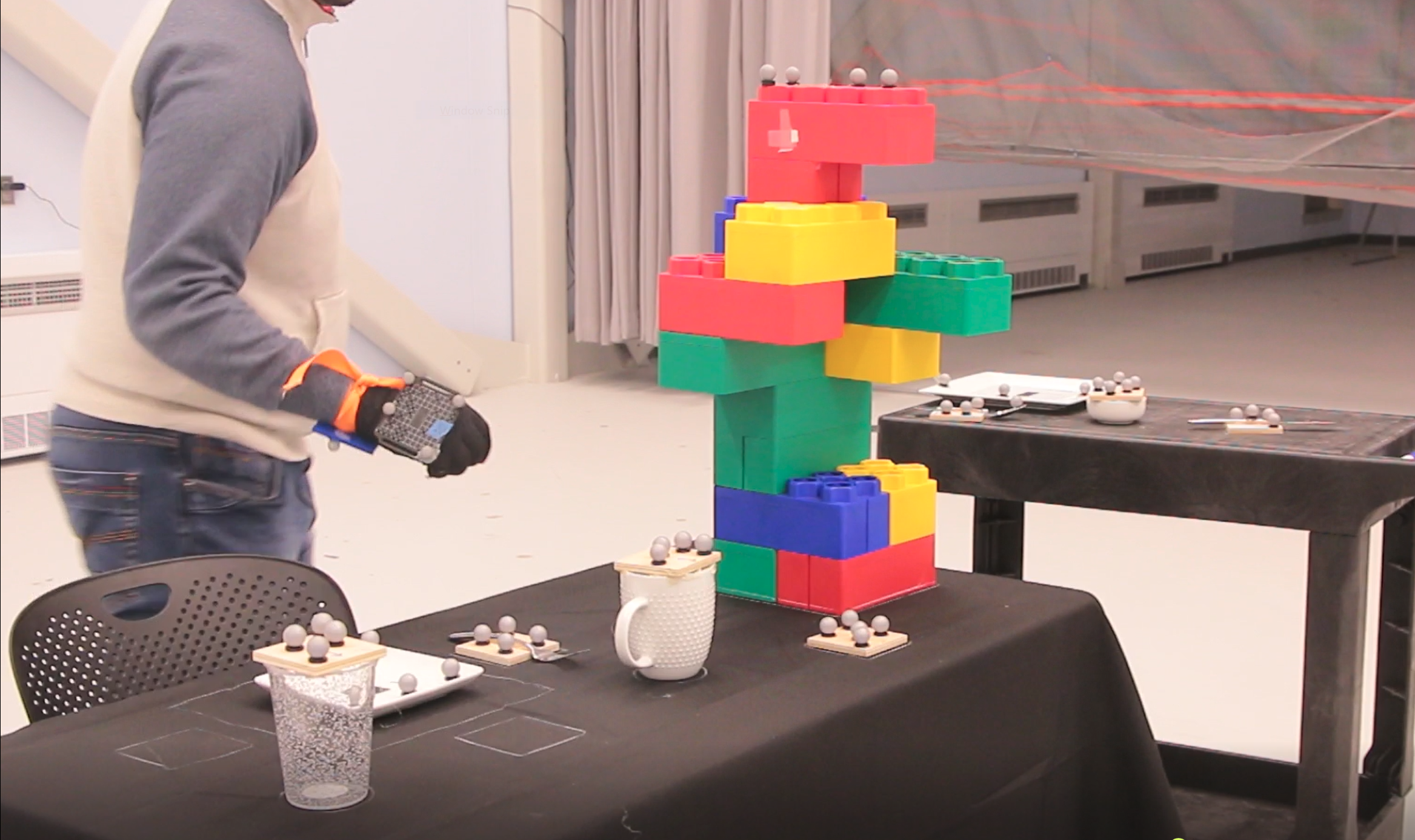}
        \caption{}

        \label{fig:DinnerB}
    \end{subfigure}
    %\vskip\baselineskip
    \caption{Figure \ref{fig:DinnerA} depicts all the final configurations. Figure \ref{fig:DinnerB} depicts the demonstration setup. (Photographed by the authors in April 2017.)}
    \label{fig:DinnerTable}
    \vspace{-15pt}
\end{figure*}

\begin{figure}
    \centering

    \begin{subfigure}[b]{0.3\textwidth}
        \centering
        \includegraphics[width=\textwidth]{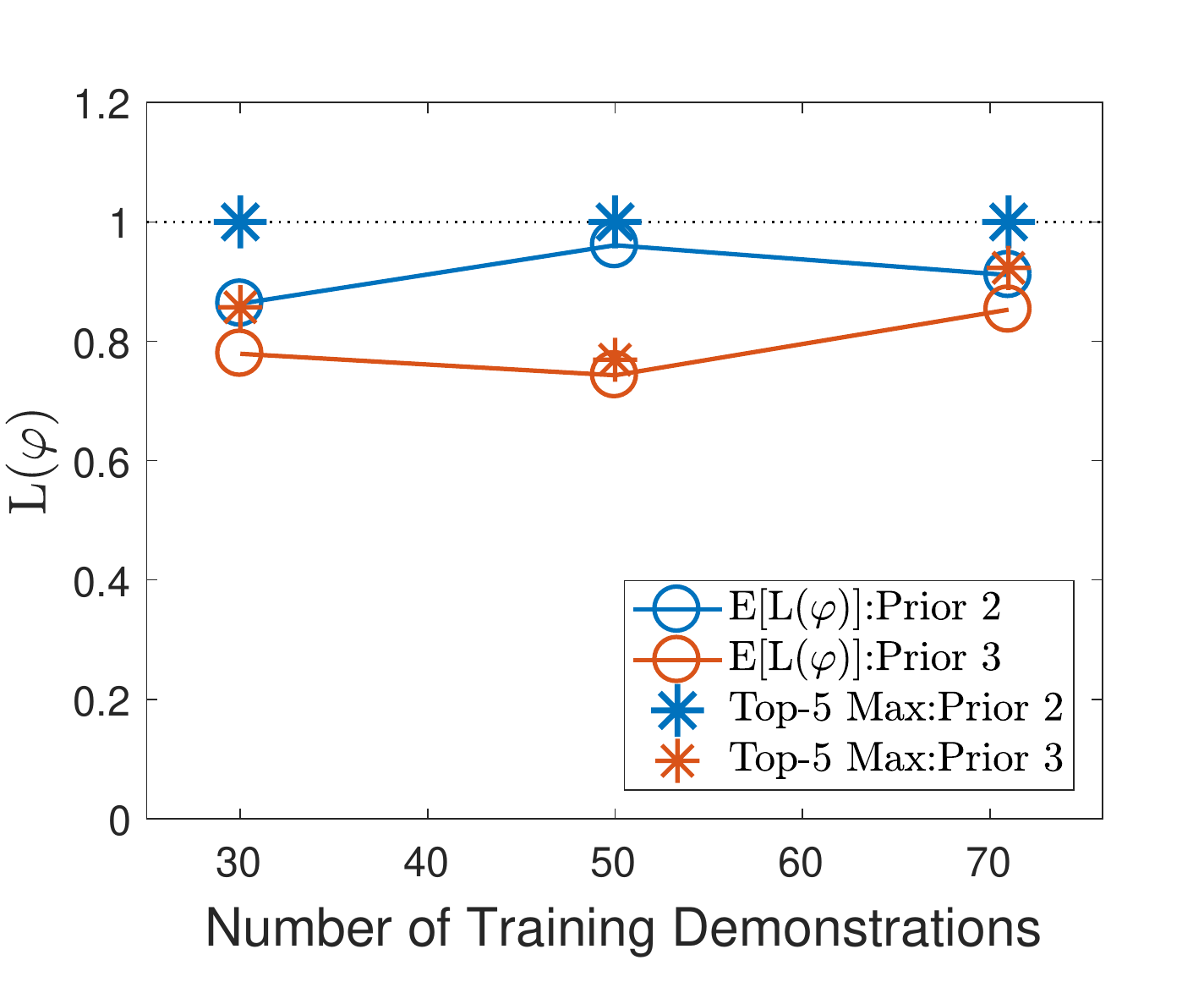}
        \caption{Dinner table $L(\varphi)$}

        \label{fig:Dinner1}
    \end{subfigure}
    \begin{subfigure}[b]{0.3\textwidth}
        \centering
        \includegraphics[width=\textwidth]{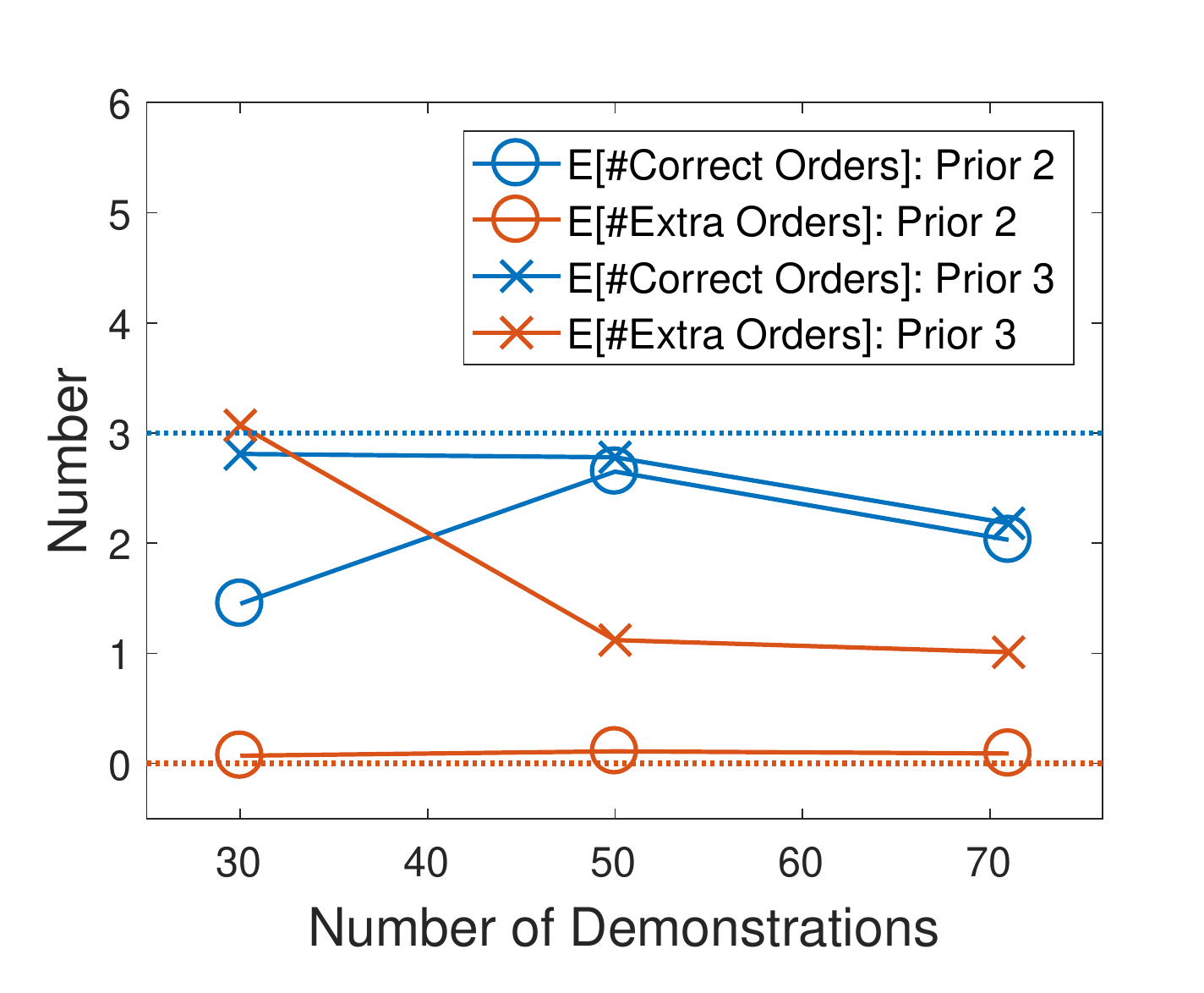}
        \caption{Dinner Table orders}

        \label{fig:Dinner2}
    \end{subfigure}
    %\vskip\baselineskip
    \caption{\autoref{fig:Dinner1} depicts the $L(\varphi)$ values for the dinner table domain, with the dotted line representing the maximum possible value. \autoref{fig:Dinner2} depicts the correct and extra orderings inferred within this domain; the dotted lines represent the number of orderings in the true specification. }
    \label{fig:Res2}
    \vspace{-10pt}
\end{figure}

We also tested our model on a real-world task: setting a dinner table. This task featured eight dining set pieces that had to be organized on a table while the demonstrator avoided contact with a centerpiece. \autoref{fig:DinnerA} depicts each of the final configurations of the dining set pieces, depending upon the type of food served. The pieces placed on the table were varied for each of the eight configurations; however, the positions of the pieces remained constant across all final configurations. A total of 71 demonstrations were collected, with six participants providing multiple demonstrations for each of the four configurations.

The eight dinner set pieces included a large dinner plate, a smaller appetizer plate, a bowl, a fork, a knife, a spoon, a water glass, and a mug; the set of pieces is represented by $\bm{\Omega}$. Each piece was tracked with a motion-capture system over the course of the demonstration, with the pose of an object $i\in\bm{\Omega}$ in the world frame represented by $\bm{T}^O_i$. In addition, the pose of the wrists of the demonstrators $\bm{T}^O_{h1}$ and $\bm{T}^O_{h2}$ were also tracked throughout the demonstration. We defined propositions that tracked whether an object was in its correct position or whether a demonstrator's wrist was too close to the centerpiece using task-space region (TSR) constraints proposed by \cite{berenson2011task}.

The origin for each TSR constraint is located at the desired final position of each object. The pose $\bm{T}^O_{w_i}$ represents the transform between the origin frame and the TSR frame for the object, $i$. The bounds for $\bm{B}_i$ represent the translation and rotational tolerances of the constraint. Finally, $\bm{P}_i$ represents the set of poses in the TSR frame that fall within the tolerance bounds. The pose of object $i$ with respect to the TSR frame is given by $\bm{T}^{w_i}_{i} = (\bm{T}^O_{w_i})^{-1}\bm{T}^O_i$. A proposition $\omega_i$ is associated with object $i$ as follows:

\begin{equation}
  \omega_i = \begin{cases}
                \text{true}, & \bm{T}^{W_i}_i\in \bm{P}_i\\
                \text{false}. & \text{otherwise}
             \end{cases}
\end{equation}

Thus, the proposition $\omega_i$ evaluates as true if the pose of object $i$ satisfies the TSR constraints, and false otherwise.

A TSR constraint is also associated with the centerpiece, where $\bm{T}^O_{c}$ represents the pose of the centerpiece with respect to the world frame, and the bounds of the constraint are defined by $\bm{B}_c$, with $\bm{P}_c$ representing the set of poses that fall within the tolerances. The poses of the demonstrator’s wrists with respect to this TSR frame are given by $\bm{T}^c_{h_i} ~\text{for}~ i \in \{1,2\}$. A proposition $\tau_c$ is associated with the centerpiece, and is defined as follows:
\begin{equation}
  \tau_c = \begin{cases}
              \text{false}, & \bm{T}^c_{h1}\in\bm{P}_c \vee \bm{T}^c_{h2}\in \bm{P}_c\\
              \text{true}, & \text{otherwise}
           \end{cases}
\end{equation}

$\tau_c$ evaluates as false if either of the wrist poses falls within the TSR bounds, and evaluates as true otherwise.

Finally, condition propositions $\pi_i ~\forall~ i \in \bm{\Omega}$ encode whether the object $i$ must be placed. Their values are set prior to the demonstration and held constant for its duration. These propositions encode the fact that serving certain courses during a meal requires specific placement of certain dinner pieces.

Based on the propositions defined above and the configurations of the dinner table, the ground-truth specifications of this task are as follows: the demonstrator’s wrists should never enter the centerpiece's TSR region (global satisfaction); if $\pi_i$ is true, then the corresponding dinner piece must be placed on the table (eventual completion); and the large plate must be placed before the smaller plate, which in turn must be placed before the bowl (ordering). We constructed the posterior distributions over candidate specification using priors 2 and 3 by incorporating subsets of the training demonstrations of varying sizes, and evaluated the similarity between the inferred specifications and the ground truth using the expected and maximum values among the top 5 a posteriori candidates of the metric $L(\varphi)$.

With prior 2, our model correctly identified the ground truth as one of the top 5 a posteriori formula candidates in all cases. With prior 3, the inferred formula contained additional ordering constraints compared with the ground truth. Using all 71 demonstrations, the MAP candidate had one additional ordering constraint: that the fork be placed prior to the spoon. Upon review, it was observed that this condition was not satisfied in only four of the 71 demonstrations.

\subsection {Evaluating Large Force Exercises}

Large-force exercises (LFE) are combat flight training exercises that involve multiple aircraft groups, with each group playing a designated role in the completion of the mission. Evaluating a LFE execution is a challenging task for the mission commander. The raw state-space of the domain includes the navigation data for each aircraft involved in the scenario (up to 36 aircrafts were included in the scenarios we simulated), along with configuration settings for each of those aircrafts (weapon stores, weapon deployments, etc.) and outcomes of combat engagements that occur throughout the scenario. The mission commander must distill this time-series and evaluate the mission based on multiple output modalities. He or she must first identify the transition points between predetermined scenario phases, then evaluate the overall success of the mission’s execution in terms of a finite number of predetermined objectives. Evaluation of the mission objectives depends not only upon the final state of the scenario, but also on the behavior of the aircrafts throughout the mission, thus making LTL a suitable grammar for representing mission objective specifications.

We evaluated the capabilities of our model to infer LTL specifications that match a mission commander’s evaluations of mission objective completion. In this section, we begin by describing the nature of the offensive counter air (OCA) mission that serves as the subject of our study. Next, we describe how these missions are evaluated by experts, and how the stated mission objectives are well-suited for use with the temporal behavioral templates we use in our candidate formulas. Finally, we describe the results obtained when applying our model to the LFE domain dataset.

\subsubsection{LFE Scenario description}
\begin{figure}
  \centering
  \includegraphics[width = 0.4\textwidth]{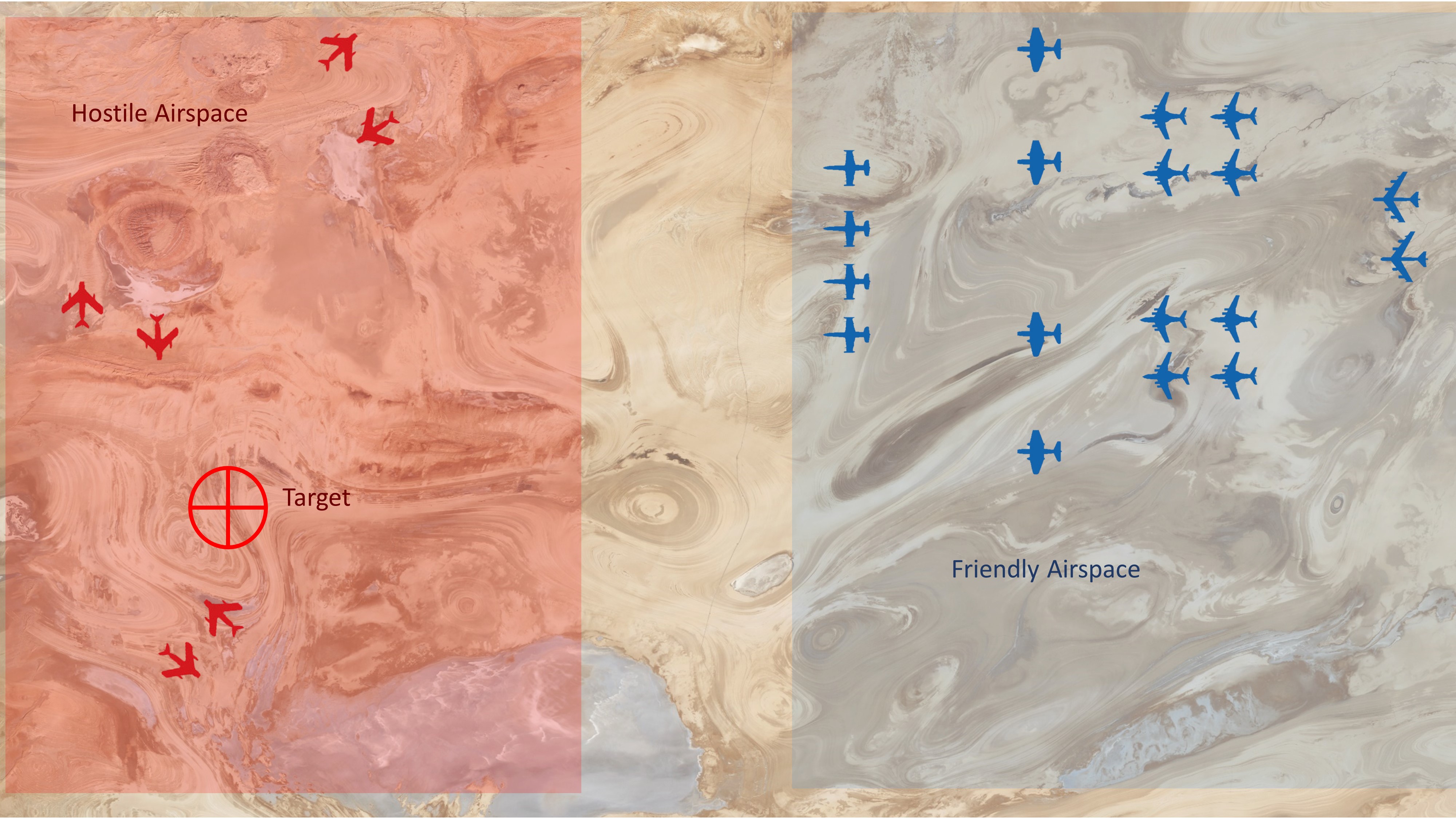}
  \caption{The starting configuration of a large-force exercise scenario. The red aircrafts are the hostile forces, and the blue are friendly forces.}
  \label{fig:lfe}
\end{figure}

Each LFE for the OCA mission we modeled consists of 18 friendly aircrafts and a variable number of enemy aircrafts and ground-based threats. Among the friendly aircrafts, there are eight escort aircrafts that are capable air-to-air fighters, eight SEAD (suppression of enemy air defenses) aircrafts capable of attacking ground-based threats, and two strike aircrafts that carry the ammunition that must be deployed in order to attack a designated ground target within a time-on-target (TOT) window. The aircrafts’ starting positions during a typical scenario are depicted in \autoref{fig:lfe}.The role of the mission commander is to debrief the participants once a LFE scenario execution is completed. During debriefing, the LFE-OCA scenario is segmented into four phases by design as follows:
\begin{itemize}
  \item{Escort Push}
  \item{Strikers Push}
  \item{Time-On-Target (TOT)}
  \item{Egress}
\end{itemize}
The mission commander must identify the times that correspond to the transitions between these mission phases, and also provide an assessment of whether the following three mission objectives were achieved:

\begin{itemize}
  \item{\textbf{MO1:}} Gain and maintain air superiority.
  \item{\textbf{MO2:}} Destroy an assigned target within the TOT window.
  \item{\textbf{MO3:}} Friendly attrition should not exceed 25\%.
\end{itemize}

Each of the mission objectives is a Boolean-valued function of the raw state-space of the LFE scenario, and the mapping between them is not explicitly known. Inputs from subject matter experts (SMEs) were also utilized to represent the mission execution in terms of certain Boolean propositions over which we can apply our probabilistic model. The propositions were defined as follows:

\begin{enumerate}
  \item{Enemy aircraft attrition (50\%, 75\%, 100\%) (three propositions).} \label{prop:hostileAttr}
  \item{Either strike aircraft fired upon.} \label{prop:strikerShotUpon}
  \item{Either strike aircraft shot down.} \label{prop:strikerShotDown}
  \item{Last munition released by strikers.} \label{prop:weaponDeployed}
  \item{Strike aircrafts flying in on-target flight phase.} \label{prop:strikerOnTarget}
  \item{Assigned target hit.} \label{prop:Target}
  \item{Friendly aircraft attrition (25\%, 50\%, 75\%)} (three propositions, each turn false if the corresponding attrition is reached). \label{prop:friendlyAttr}
\end{enumerate}

In order to generate realistic demonstrations of how the different executions unfold, the scenarios were defined in Joint Semi-Automated Forces (JSAF) -- a constructive environment capable of simulating realistic aircraft behavior. The data collected for each demonstration included the position, speed, attitude, and rates of each of the aircrafts (both friendly and hostile); the individual mission phase of each aircraft (a discrete set of phases by which the aircraft specific mission timeline can be labeled); and the firing times, designated targets, detonation times, and outcomes of each weapon deployment over the course of the scenario. The mapping from the collected data to the Boolean propositions stated above is well defined.

In order to apply our probabilistic model to the LFE domain, we defined the sets $\bm{\tau}$ and $\bm{\Omega}$. The propositions \ref{prop:friendlyAttr}, \ref{prop:strikerShotUpon}, and \ref{prop:strikerShotDown} were included in the set $\bm{\tau}$ as candidates for global satisfaction. The propositions \ref{prop:hostileAttr}, \ref{prop:weaponDeployed}, \ref{prop:strikerOnTarget}, and \ref{prop:Target} were included in $\bm{\Omega}$ as candidates for eventual completion.

\subsubsection{Data collection}
A total of 24 instances of LFEs were simulated and included in the dataset. Each instance had a different outcome with respect to the mission objectives, based on the different outcomes of engagements between friendly and hostile forces. Each scenario was evaluated by an SME acting as a mission commander performing a manual debrief. The primary annotation task was to evaluate whether each of the objectives was successfully achieved upon mission completion. The secondary annotation task was to determine the segmentation points among the four scenario phases on the mission timeline. The segmentation task is not directly relevant to specification inference, but we used the labels to simultaneously train a secondary classifier in one of the baselines.

\subsubsection{Benchmarks}
\label{ss:benchmarks}

 The training data for evaluations of LFEs consists of both acceptable and unacceptable demonstrations, along with the label for that demonstration; thus, it can be viewed as a supervised learning problem. We decided to compare the classification accuracy of our model against a classifier trained with a recurrent neural network as the underlying architecture.

\begin{enumerate}
  \item{\textbf{Stand-alone:}} Here, the recurrent neural network is trained to jointly optimize the binary cross-entropy for classification of each of the three mission objectives. The loss functions for all the mission objectives are equally weighted. The recurrent neural networks are composed of long and short-term memory (LSTM) modules (\cite{Hochreiter}), along with their bidirectional variants (\cite{graves2005bidirectional}). Such models have shown state-of-the-art performance during time-series classification tasks (\cite{RNN2016}). These models -- henceforth referred to as `LSTM' and `Bi-LSTM,' respectively -- were trained using only the time-series of the propositions as inputs.

  \item{\textbf{Coupled:}} In prior research, performance improvements on a primary task have been observed due to simultaneous training on a secondary related task (\cite{NIPS2015_5775}). We hypothesized that simultaneously training the classifier on the secondary task of identifying scenario phases might improve classification accuracy compared with a standalone RNN. The loss functions used were binary cross-entropy for each of the mission objectives and categorical cross-entropy for the scenario phase identification. The overall loss function was an equally weighted sum of the individual cost functions. These models were also composed of LSTM modules and their bidirectional counterparts, and are referred to as `LSTM Coupled' and `Bi-LSTM Coupled,' respectively. These models were trained using the propositions and collected flight phase data.
\end{enumerate}

\subsubsection{Evaluations}
\label{ss:evaluation}
The classification models were evaluated through a four-fold cross-validation wherein the training dataset was divided into four equal partitions, with three of the partitions used for training (18 scenarios) and testing performed on the remaining partition (6 scenarios); this was repeated across all partitions. The predictions of the model for each of the scenarios were assimilated at the end. We also applied our model to the entire dataset in order to analyze which of the propositions were included in the maximum a posteriori estimate of the specifications. The overall accuracy of the classifiers was evaluated using the F1 score on all the predictions for both the possible outcomes of the mission objectives ('Achieved' and 'Failed') for each mission objective.

\subsubsection{Results}
\label{ss:lferesults}

\begin{table}
  \footnotesize
  \caption{Weighted F1 scores for both scenario outcomes for each of the classifiers.}
  \label{Tab:LFEResults}
  \centering
  \begin{tabular}{llll}
    \toprule
    \textbf{Classifier} & \textbf{MO1} & \textbf{MO2} & \textbf{MO3}                   \\
    \midrule

    LSTM            &0.533     &0.533   &0.481   \\
    Bi-LSTM         &0.533     &0.533   &0.481   \\
    LSTM Coupled    &0.533     &0.533   &0.481   \\
    Bi-LSTM Coupled &0.533     &0.533   &0.481   \\
    BSI (Prior 2)   &\textbf{0.674}     &\textbf{0.712}   &\textbf{0.877}   \\
    BSI (Prior 3)   &\textbf{0.674}     &0.676   &\textbf{0.877}   \\
    \bottomrule
  \end{tabular}
\end{table}

As presented in \autoref{Tab:LFEResults}, our model outperformed RNN-based supervised learning models. With a four-fold split of training and test data, prior 2 seemed to outperform prior 3; one possible explanation would be that the bias of prior 3 toward longer task chains might result in a higher rate of false negatives.

We also noticed the tendency of RNN models to collapse to predicting the most commonly occurring outcome in the training set for all values of inputs. Thus, the model was unable to achieve high accuracies even on the training set, suggesting that it is not only the small size of the dataset that results in poor performance. This might indicate that either greater model capacity or a different model architecture may be required.

Next, we analyzed the maximum a posteriori formula returned by our model using prior 2, and the F1 scores obtained were 0.959, 0.918, and 0.959 for the three mission objectives, respectively. The compositional structure of the model allowed us to examine the propositions included in the formulas and interpret the decision boundaries of the classifiers; the results were as follows:

\begin{enumerate}
  \item{\textbf{MO1 (Gain and maintain air-superiority)}} The propositions included in $\varphi_{global}$ were \ref{prop:friendlyAttr}, \ref{prop:strikerShotDown}, and \ref{prop:strikerShotUpon}; these correspond to a maximum allowable friendly attrition rate of less than 25\%, and enforcing the condition that the strikers were never fired upon or shot down, respectively. (This is consistent with the definition of air superiority.) The propositions included in $\varphi_{eventual}$ were \ref{prop:weaponDeployed}, \ref{prop:hostileAttr}, and \ref{prop:strikerOnTarget}; these correspond to strikers eventually releasing their weapons, the friendly forces shooting down 75\% of the enemy fighters, and strike aircrafts eventually reaching their on-target flight phase, respectively. (Again, the included propositions indicate that gaining air superiority allowed strikers to operate freely.) Finally, $\varphi_{order}$ enforced that friendly forces shot down 50\% of the hostile air threats before strikers released their weapons.

  \item{\textbf{MO2: (Destroy assigned target)}} The propositions included in $\varphi_{global}$ were \ref{prop:friendlyAttr} and \ref{prop:strikerShotDown}; these represent a maximum friendly attrition of 50\%, and only enforcing that the strikers were never shot down, respectively. (Note that this does not enforce the condition that strikers were never fired upon.) $\varphi_{eventual}$ included \ref{prop:hostileAttr}, \ref{prop:weaponDeployed}, \ref{prop:strikerOnTarget}, and \ref{prop:Target}; these represent eventually shooting down all hostile aircrafts (which would seem unnecessary), strikers entering their on-target flight phase, eventually releasing their weapons — and, most importantly, attacking the assigned target. $\varphi_{order}$ enforced the condition that the friendly aircrafts had to shoot down all hostiles before the close of the TOT window.

  \item{\textbf{MO3: No more than 25\% friendly losses:}} The propositions in $\varphi_{global}$ included \ref{prop:friendlyAttr}, \ref{prop:strikerShotUpon}, and \ref{prop:strikerShotDown}; these correctly enforced that no more than 25\% friendly aircrafts could be shot down, and also that the strikers were never shot down or fired upon. $\varphi_{eventual}$ included \ref{prop:hostileAttr}, \ref{prop:weaponDeployed}, and \ref{prop:strikerOnTarget}, representing 75\% hostile force attrition, and enforced that the strikers had to eventually enter their on-target phase and deploy their weapons. No orders were included in the formula. The propositions that enforced weapon deployment by strikers and requisite hostile attrition were not required for this objective to be fulfilled; however, they were included by the model due to their frequent occurrence with objective completion. The compositional nature of the model allows the user to identify constraints that will be easily enforced.

\end{enumerate}

\section{Conclusion}
\label{sec:conclusion}
In this work, we presented a probabilistic model to infer task specifications in terms of three behaviors encoded as LTL templates. We presented three prior distributions that allow for efficient sampling of candidate formulas as per the templates. We also presented a likelihood function that depends only upon the number of conjunctive clauses in the candidate formula, and is transferable across domains as it requires no information about the domain itself. Finally, we demonstrated our model on three distinct evaluation domains. On the domains where the ground-truth specifications were known, we demonstrated the capability of our model to identify the ground-truth  specification with up to 90\% similarity , in both a low-dimensional synthetic domain and a real-world dinner table domain. In the large-force exercise domain, where the ground-truth specifications are not known, we showed the ability of our model to align its predictions with those of an expert to a greater extent than supervised learning techniques. We also demonstrated our model’s ability to explain its decision boundaries due to the compositional nature of the formula template.

\section*{Acknowledgements}
This work has been funded by the Lockheed Martin corporation and the Air Force Research Laboratory. We would like to acknowledge Dr. Kevin Gluck, and Dr. Donald Duckro at the Airman Systems Directorate and Zachary 'Zen' Wallace at the Rickard Consulting Group for their expertise in mission design and analysis that were instrumental in creating the LFE scenario. We would also like to thank David Macannuco at the Lockheed Martin corporation for his expertise in modeling and simulation.

\bibliographystyle{SageH}
%%Vancouver (numbered)
%\bibliographystyle{SageV}
\bibliography{refs}

\end{document}